\pdfoutput=1

\documentclass[11pt]{article}

\usepackage{ACL2023}

\usepackage{times}
\usepackage{latexsym}

\usepackage[T1]{fontenc}

\usepackage[utf8]{inputenc}

\usepackage{microtype}

\usepackage{inconsolata}

\usepackage{microtype}
\usepackage{graphicx} 
\usepackage{float} 
\usepackage{subfigure} 
\usepackage{wrapfig}
\usepackage{graphicx}
\usepackage{amsfonts}
\usepackage{bm}
\usepackage{array,amsmath}
\usepackage{amssymb}
\usepackage{dsfont}
\usepackage{float}
\usepackage{algorithm}
\usepackage{algorithmicx}
\usepackage{algpseudocode}
\usepackage{pgf,tikz}
\usepackage{mathrsfs}
\usepackage{multirow}
\usepackage{booktabs}
\usetikzlibrary{arrows}
\usepackage{threeparttable}
\usepackage[inline]{enumitem}
\usepackage{multicol}

\def\figref#1{Figure~\ref{fig:#1}}
\def\figlabel#1{\label{fig:#1}\label{p:#1}}

\def\tabref#1{Table~\ref{tab:#1}}

\def\tablabel#1{\label{tab:#1}\label{p:#1}}

\def\secref#1{\S\ref{sec:#1}}
\def\seclabel#1{\label{sec:#1}}

\newcounter{notecounter}

\newcommand{\enoteson}{\long\gdef\enote##1##2{{
			\stepcounter{notecounter}
			{\large\bf \hspace{1cm}\arabic{notecounter} $<<<$ ##1: ##2 $>>>$\hspace{1cm}}}}}
\enoteson
\usepackage{xcolor}
\usepackage{soul}
\usepackage{xspace}

\usepackage{color}

\def\themethod{PARC\xspace}

\long\def\eat#1{}

\newcommand\blfootnote[1]{%
  \begingroup
  \renewcommand\thefootnote{}\footnote{#1}%
  \addtocounter{footnote}{-1}%
  \endgroup
}

\title{Cross-Lingual Retrieval Augmented Prompt for Low-Resource Languages}

\author{Ercong Nie$^{\star}$ $^{1,2}$ \qquad Sheng Liang$^{\star}$ $^{1,2}$ \qquad Helmut Schmid$^{1}$ \qquad Hinrich Sch\"utze$^{1,2}$ \\
$^{1}$Center for Information and Language Processing (CIS), LMU Munich, Germany \\
$^{2}$ Munich Center for Machine Learning (MCML), Germany \\
\texttt{\{nie, shengliang\}@cis.lmu.de}}

\begin{document}
\maketitle

\begin{abstract}
Multilingual Pretrained Language Models (MPLMs) 
perform strongly in
cross-lingual transfer. We propose
\textbf{P}rompts \textbf{A}ugmented by
\textbf{R}etrieval \textbf{C}rosslingually
(\textbf{\themethod}) to improve  zero-shot
performance on low-resource languages (LRLs) by augmenting
the context with prompts consisting of semantically similar sentences retrieved
from a high-resource language (HRL). \themethod
improves zero-shot performance on three downstream tasks
(sentiment classification, topic categorization,
natural language inference) with multilingual parallel test
sets across 10 LRLs covering 6 language families in 
unlabeled (+5.1\%) and labeled settings
(+16.3\%). \themethod also outperforms
finetuning by 3.7\%.  We find a significant
positive correlation between cross-lingual transfer
performance on one side, and the similarity between 
high- and low-resource languages as well as the amount of
low-resource pretraining data on the other side. A
robustness analysis suggests that \themethod has the
potential to achieve even stronger performance with more
powerful MPLMs. 
The codes and data for our work are available: \url{https://github.com/ercong21/parc}
\blfootnote{$^\star$ Equal Contribution.}

\end{abstract}


\section{Introduction}

\begin{figure}[!t]
	\centering  
	\subfigure[Retrieval from high-resource language corpora]{  
		\begin{minipage}{7cm}
			\centering    
			\includegraphics[scale=0.434]{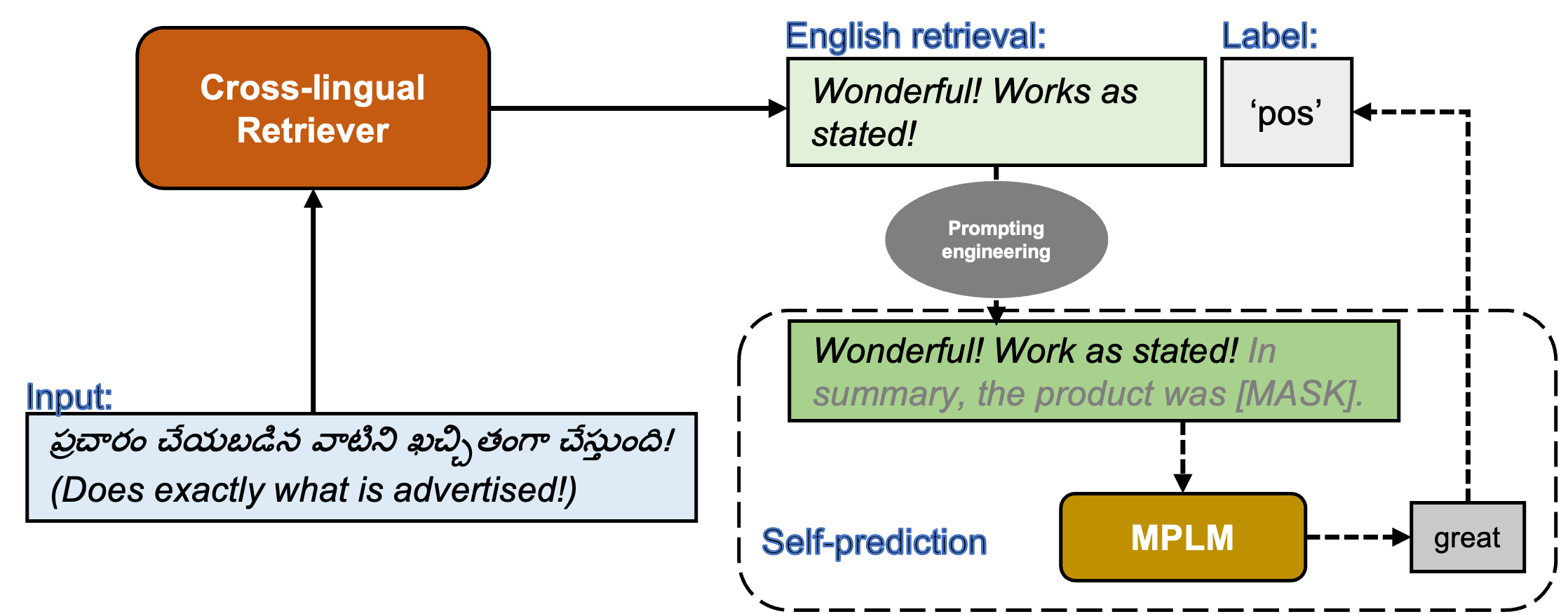} 
		\end{minipage}
	}
	\subfigure[Prediction with a retrieval-augmented prompt]{ 
		\begin{minipage}{7cm}
			\centering    
			\includegraphics[scale=0.465]{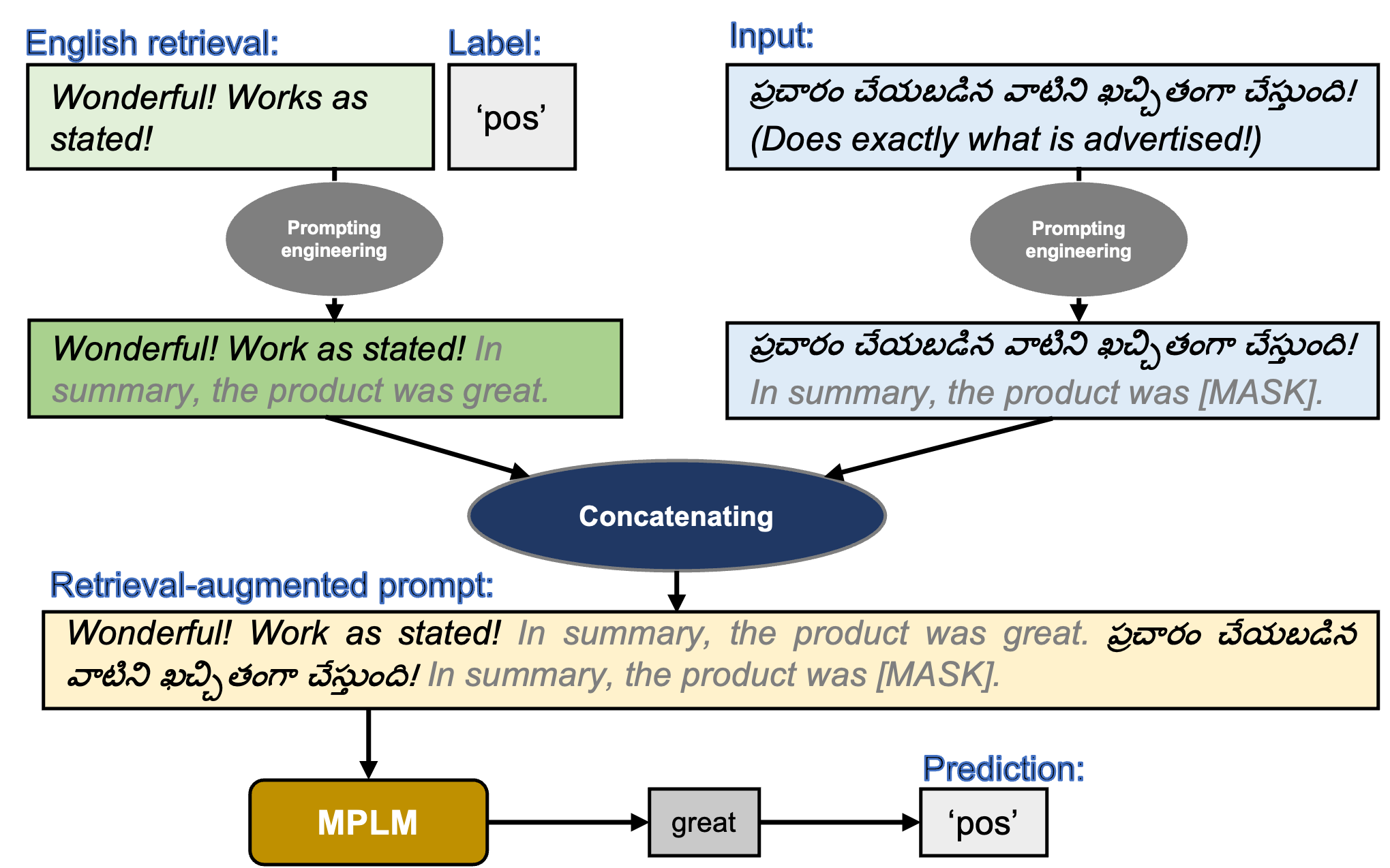}
		\end{minipage}
	}

	\caption{Main idea of \themethod: we enhance
	zero-shot learning for low-resource languages (LRLs)
	by cross-lingual retrieval
	from \textbf{labeled}/\textbf{unlabeled}
	high-resource languages (HRLs). (a) An LRL input
	sample is taken as query by the cross-lingual
	retriever to retrieve the semantically most similar
	HRL sample from the HRL corpus. The label of the
	retrieved HRL sample is obtained either from the
	corpus (\textbf{labeled} setting) or by
	self-prediction (\textbf{unlabeled} setting). (b)
	The retrieved HRL sample together with its label and
	the input sample are reformulated as prompts. The
	cross-lingual retrieval-augmented prompt is created
	by concatenation and taken by the MPLM for
	prediction. Our experiments show that \themethod
	outperforms other zero-shot methods and even finetuning.}    
	\figlabel{pipeline}   

\end{figure}

Multilingual pretrained language models (MPLMs) \citep{devlin-etal-2019-bert,conneau-etal-2020-unsupervised,liu-etal-2020-multilingual-denoising,xue-etal-2021-mt5,shliazhko2022mgpt}, pretrained on multilingual corpora with $>$100 languages,
exhibit strong multilinguality
on downstream tasks \citep{hu2020xtreme}. 

Low-resource languages (LRLs), for which little text data is available for pretraining monolingual pretrained language models (PLMs), benefit from MPLMs.
However, the lack of LRL data leads to an imbalanced language distribution
in the pretraining corpora of MPLMs \citep{wu-dredze-2020-languages}. 
LRLs are therefore under-represented in pretraining, resulting in bad performance.
Furthermore, the scarcity of domain- or task-specific annotated data of LRLs
makes it difficult to apply the pretraining-finetuning paradigm
to LRLs \citep{lauscher2020zero}. 
Given that the pretraining-finetuning paradigm always has a high demand for domain-specific labeled data, 
another line of research -- prompt-based learning -- emerges, focusing on exploiting large pretrained language models by reformulating the input.
The prompt is designed to help PLMs ``understand'' the task better
and ``recall'' what has been learned during the pretraining. 
In particular, \citet{brown2020language} propose 
a simple in-context learning approach 
without any finetuning,
which adds training examples 
as additional context to test examples.
Instead of using random examples as context, KATE \citep{liu-etal-2022-makes} and SOUP \citep{liu2022semantic} retrieve semantically similar examples as prompt 
for monolingual in-context learning.
The above mentioned prompt-based learning techniques require no parameter updating, while there
is also work employing sampled similar examples for prompt-based funetuning \citep{gao-etal-2021-making}.
Unlike \citet{brown2020language} who created prompts with manually selected examples,
these approaches augment the context by retrieving related information from external corpora, allowing the PLMs to capture more domain- or task-specific knowledge.
The prompt-based method offers a new form of zero-shot or
few-shot learning in multilingual NLP studies. It involves performing a specific task using prompts, without labeled data in the target language and has the potential of being an effective method
for LRLs lacking annotated data.

Our work improves 
 the zero-shot transfer learning performance of LRLs on three different classification tasks by taking advantage of cross-lingual information retrieval and the multilinguality of MPLMs.
Specifically, we retrieve semantically similar cross-lingual sentences as prompts and use the cross-lingual retrieval information to benefit the LRLs from the multilinguality of MPLMs and 
achieve better performance in the zero-shot setting\footnote{Different from the zero-shot cross-lingual transfer learning where MPLMs are finetuned on HRLs \citep{hu2020xtreme}, our zero-shot setting does not involve finetuning. Details in \secref{zeroshot_setting}}.
Our main contributions are: (1) We 
propose \textbf{P}rompts \textbf{A}ugmented by \textbf{R}etrieval \textbf{C}rosslingually (\textbf{\themethod}), a pipeline for integrating retrieved cross-lingual information into prompt engineering for zero-shot learning (\figref{pipeline}). 
(2) We conduct experiments on several multilingual tasks, 
showing that \themethod improves the zero-shot performance on LRLs
by retrieving examples from both labeled and unlabeled HRL corpora.
(3) To find an optimal 
configuration of our \themethod pipeline, 
we
conduct 
a comprehensive study on the variables 
that affect the zero-shot performance: the number of prompts, the choice of HRL, and the robustness w.r.t.\ other retrieval methods and MPLMs.

\section{Related Work}

\paragraph{Retrieval methods} External knowledge extracted by information retrieval is often leveraged to solve NLP tasks. Two types of representations have been used for retrieval: (1) sparse bag-of-words representations \citep{chen-etal-2017-reading, Wang2018R3RR}, and (2) dense representation learned by neural networks \citep{Qu2020RocketQAAO}. Dense representations come either from contextual token embeddings \citep{May2019OnMS, Zhang2020BERTScoreET} or from sentence encoders \citep{conneau-etal-2017-supervised, cer2018universal}. \citet{reimers-2019-sentence-bert} propose sentence transformers to create semantically meaningful sentence embeddings by applying siamese and triplet network structures to transformer-based pretrained language models. By using knowledge distillation, sentence transformers can be expanded to support various languages as multilingual sentence transformers \citep{reimers-2020-multilingual-sentence-bert}, allowing for cross-lingual retrieval.

\paragraph{Retrieval augmented prompt} \citet{brown2020language} show that large-scale pretrained language models such as GPT-3 can learn to perform a task by putting examples of input-output pairs into the input as context. The in-context learning method simply concatenates the input with examples randomly extracted from the training set. Recent studies \citep{gao-etal-2021-making, liu-etal-2022-makes, liu2022semantic} augment the prompts for pre-trained models by sampling semantically similar examples. They apply the retrieval augmented method to discrete prompts, which are represented by tokens instead of vectors in a continuous space.
They 
use them either for finetuning in few-shot settings or for zero-shot learning. \citet{Chowdhury2022NoveltyCP} use a similar kNN-based retrieval method for tuning the soft prompts in a continuous space with a standard supervised training setup. Previous work focused on monolingual retrieval-augmented prompts. Our work applies cross-lingual retrieval to discrete prompts in a scenario without parameter updating. To the best of our knowledge, our work is the first to investigate prompt learning augmented by cross-lingual retrieval. 

\paragraph{Multilingual prompt learning} Despite the success of prompting in English, prompting in multilingual tasks has not been extensively studied. \citet{winata-etal-2021-language} show the multilingual skills of LMs mainly trained on English data in prompt learning by giving them a few English examples as context but testing them on non-English data. Some recent works investigate the prompt learning with multilingual PLMs \citep{zhao-schutze-2021-discrete, Huang2022ZeroshotCT}. Unlike our work, they focus on finetuning or prompt tuning requiring parameter updating.
We apply our method to 
LRLs in a zero-shot setting without adjusting the model parameters.
	
\section{Methodology}
\seclabel{Methodology}


This work aims to improve the performance of MPLMs on LRLs in the zero-shot setting
by leveraging retrieved cross-lingual contents from HRLs. For that, we design the \themethod pipeline that can be applied to labeled and unlabeled scenarios, i.e., the HRL information can be retrieved from either labeled or unlabeled corpora.


As \figref{pipeline} shows, the \themethod pipeline consists of two steps: (a) Cross-lingual retrieval from high-resource language corpora, and (b) prediction with a retrieval-augmented prompt. \figref{pipeline} shows an example: A Telugu input sentence from a sentiment classification task is firstly fed into the cross-lingual retriever to fetch the semantically closest sample from the HRL corpus, i.e. English in this case. In the second step, the retrieved HRL sample together with its label and the LRL input sentence are transformed into a prompt. For prompt-based classification, we need (i) a \textit{pattern} which converts the input sentence into a cloze-style question with a mask token, and (ii) a representative word (called \textit{verbalizer}) for each possible class.
Converting the classification task into a cloze-style question aligns seamlessly with the framework of our proposed \themethod method, because it not only performs zero-shot learning well but, more significantly, facilitates better integration of the retrieved cross-lingual contexts.

In our example, we use the pattern  $P(X)=X \circ$ ``\texttt{In summary, the product was [MASK].}'' to convert the retrieved English sentence into ``\texttt{Wonderful! Works as stated! In summary, the product was [MASK].}'', where $\circ$ is the string concatenation operator. 
A verbalizer such as \{\texttt{pos} $\rightarrow$ ``great'', \texttt{neg} $\rightarrow$ ``terrible''\}, which maps the original labels \{\texttt{pos}, \texttt{neg}\} onto words in the vocabulary, is then used to replace the \texttt{[MASK]} token  with the verbalized label word ``great'', standing for the correct label \texttt{pos} of this sentence. We call the resulting English sentence (in our example: ``\texttt{Wonderful! Works as stated! In summary, the product was great.}'') the ``cross-lingual context''. 
At last, we fill the same pattern with the input Telugu sentence and append it to the cross-lingual context. We feed this cross-lingual retrieval augmented input to the MPLM. The MPLM returns for each of the verbalizers its probability of being the masked token. 

More formally, let $X^L_i \in D^L$ be the input sample from the LRL test set, $(X^H_j, y_j) \in D^H_{lb}$ and $X^H_j \in D^H_{un}$ denote the HRL data from the \emph{labeled} and \emph{unlabeled} corpora, respectively, where $X_j$ is the text sample and $y_j$ its class label from a label set $Y$. 
As Eq. \eqref{retrieve} shows, the cross-lingual retriever $CLR$ takes the HRL corpora $D^H$ and a given LRL input sentence $X^L_i$. It returns an ordered list of HRL sentences $D^{R_i}$ according to the semantic similarity. We then have $(X^{R_i}_k, y^{R_i}_k) \in D^{R_i}_{lb}$ and $X^{R_i}_k \in D^{R_i}_{un}$ for labeled and unlabeled scenarios, respectively, where $X^{R_i}_k$ is the $k$-th most similar HRL sentence to the LRL input $X^L_i$. 
\begin{equation}
    D^{R_i} = CLR(X^L_i, D^H)
    \label{retrieve}
\end{equation}
The prompt pattern $P(.)$ converts an HRL input sentence $X^{R_i}_k$ into a cloze-style form with a mask token. The verbalizer $v(.)$ is a bijective mapping from the set of class labels $Y$ to a set of verbalized words $V$ from the HRL vocabulary. 
We use the verbalized label word to fill in the mask token in the prompt pattern, and construct the cross-lingual context $C^i_k$ for the input $X^L_i$ with the $k$-th most similar HRL sample $X^{R_i}_k$:
\begin{equation}
\begin{aligned}
    C^i_k = P(X^{R_i}_k, v(y^{R_i}_k)) \label{context}
\end{aligned}
\end{equation}

The cross-lingual context $C^i_k$ is then concatenated with the prompted LRL input as the input $I$ to the MPLM:
\begin{equation}
    I_i = C^i_k \circ P(X^L_i)
    \label{concat}
\end{equation}

The MPLM $M$ performs masked token prediction and returns the probabilities $p = M(I_i)$ of all candidate words for the masked token in $I_i$. We predict the class $\hat{y}$ whose verbalizer $v(\hat{y})$ received the highest probability from model $M$:
\begin{equation}
    \hat{y} = \arg\max_{y\in Y} p(v(y))
    \label{argmax}
\end{equation}

In the labeled scenario, we obtain the correct label $y^{R_i}_k$ of the HRL
sentence from $D^{R_i}_{lb}$.  In the unlabeled scenario, we predict the
label using the same prompt-based classification method without
cross-lingual context, similar to Eq. \eqref{argmax}. 
We call this the \emph{self-prediction} method:
\begin{equation}
    \hat{y}^{R_i}_k = \arg\max_{y\in Y} M(P(X^{R_i}_k), v(y))
    \label{self_pred}
\end{equation}

In order to use more cross-lingual information, we retrieve the $K$ most
similar HRL samples.
With each sample,
we obtain verbalizer probabilities as described above and retrieve the
class whose verbalizer has the largest sum of probabilities.  We call
this method the Bag-of-Retrieval (BoR) strategy. We also tried
concatenating the different cross-lingual contexts (CONC method), but
the resulting performance has been worse 
(see \tabref{agnews_conc} in the Appendix).

\section{Experimental Setup}
\seclabel{exp}

\subsection{Datasets}
\paragraph{Base Datasets}
Three representative classification tasks are selected for evaluation in this work: 
binary sentiment analysis on Amazon product reviews~\citep{marc_reviews}, 
topic classification on AG News texts~\citep{Zhang2015CharacterlevelCN}, 
and natural language inference on XNLI~\citep{Conneau2018XNLIEC}. 

\textbf{Amazon Reviews} dataset categorizes the shopping reviews into 5 star ratings from 1 to 5. 
In order to satisfy a binary classification setting, 
we select the reviews with rating 1 as \texttt{negative} (0) and 5 as \texttt{positive} (1) for our experiments. 
The following pattern $P(X)$ and verbalizer $v$ are defined for an input review text $X$:
\begin{itemize}
	\item $P(X) = X \circ \text{``All in all, it was [MASK].''}$
	\item $v(0)=\text{``terrible''}$, $v(1)=\text{``great''}$
\end{itemize}

\textbf{AG News} is a collection of more than 1 million news articles for topic classification. 
The news topic categories contained in the dataset are \texttt{World} (0), \texttt{Sports} (1) , \texttt{Business} (2), and \texttt{Tech} (3). The pattern and verbalizers are as follows:
\begin{itemize}
	\item $P(X)=\text{``[MASK] News: ''} \circ X$
	\item $v(0)=\text{``World''}$, $v(1)=\text{``Sports''},
	\\v(2)=\text{``Business''}$, $v(3)=\text{``Tech''}$
\end{itemize}

\textbf{XNLI} is a multilingual version of the MultiNLI dataset \citep{N18-1101}. We use a subset of the original XNLI dataset in our experiment. The text in each data item consists of two parts. Sentence A is the premise and sentence B is the hypothesis. 
The NLI task is to predict the type of inference between the given premise and hypothesis among the three types: \texttt{entailment} (0), \texttt{neutral} (1) and \texttt{contradiction} (2). 
For a given sentence pair $X_1$ and $X_2$, 
we design the pattern and verbalizer as:
\begin{itemize}
	\item $P(X_1, X_2) = X_1 \circ \text{``? [MASK],'' } \circ X_2$
	\item $v(0) = \text{``Yes'' }$, $v(1)=\text{``Maybe'' }, v(2)=\text{``No'' }$
\end{itemize}

\paragraph{Construction of Multilingual Parallel Test Sets} 

Parallel test datasets for evaluating cross-lingual transfer
performance on LRLs are rare. 
However, recent research conducted by \citet{hu2020xtreme, liu-etal-2022-mulzdg} shows that automatically translated test sets are useful for measuring cross-lingual performance. 
Hence, we adopt their methodology and construct datasets for different tasks by
automatically translating English test sets to targeted LRLs.
We use the Python API of the Google Translate System to implement the construction of multilingual parallel test sets in our experiment. 
We also validate the translation effectiveness and quality. 
The original XNLI datasets include two low-resource languages that are used in our experiments (Swahili and Urdu), so we use them as the ``gold'' standard for our translation validation.
We compare the cross-lingual transfer performance on translation test sets and original test sets of XNLI. We also measure the translation quality by using the original sets as gold standard. Through the validation conducted on these two languages within the XNLI task, we infer that the translation method is effective and could be generalized to other languages and tasks.
Detailed results are shown in Appendix \secref{translation}.

Following 
\citet{wu-dredze-2020-languages}, we regard languages with a WikiSize\footnote{WikiSize less than 7 means that the Wikipedia corpus of the language is smaller than 0.177 GB.} of less than 7 as LRLs. We 
construct a test set consisting of 10 LRLs in 6 language families:
Indo-European (Afrikaans - af, Urdu - ur), 
Austronesian (Javanese - jv, Tagalog - ta), 
Altaic (Mongolian - mn, Uzbek - uz), 
Dravidian (Tamil - tl and Telugu - te), 
Sino-Tibetan (Burmese - my),
and Niger-Congo (Swahili - sw). \tabref{dataset} in the Appendix shows more information on the test sets.

\paragraph{HRL Corpora}
To retrieve rich and diverse information, a large-scale general corpus
or knowledge base in the different HRLs would be
the ideal sentence retrieval pool.
In practice, however, a trade-off is necessary in order to save computational resources. 
Following \citet{wang-etal-2022-training},
we therefore use the task-specific labeled training set of English as the sentence pool in our experiments.
The selection of the HRL will be discussed in \secref{choiceoflangs}.

\subsection{Baseline}
\seclabel{baseline}
We compare \themethod with the following baselines in both labeled and unlabeled settings:

\textbf{MAJ} The majority baseline. Since we construct the test sets to be balanced, MAJ is equivalent to random guess.

\textbf{Random} We randomly retrieve a cross-lingual sentence as
prompt, similarly to the simple in-context learning using examples
without semantic similarity to the input \citep{brown2020language}.

\textbf{Direct} The pattern filled with the input sample is directly
fed to the MPLM for prediction, without adding cross-lingual context to the prompts.

\textbf{Finetune} The MPLM is first finetuned with the retrieved high resource sentences. Then the low-resource test input is predicted by the finetuned MPLM. We use the Cross Entropy Loss as the objective function for finetuning and AdamW for optimization with a learning rate of 1e-5. Since the finetuning data is very limited, we only train for a single epoch to avoid overfitting.

Our test sets are constructed by machine translation. Therefore we
cannot apply a translation baseline, where we translate the input
sample into the high resource language before feeding it to the MPLM.
The Appendix presents a different experiment where we compare with a translation baseline.

\subsection{Models}
\paragraph{Cross-Lingual Retriever}
The retrieval methods used in monolingual NLP are either based on
sparse or dense representations. Sparse representations such as
BM25 \citep{manning2008introduction} which is based on term frequency,
cannot be used for cross-lingual retrieval as the shared words across
different languages are normally scarce. Therefore dense
representations from deep learning methods such as
LASER \citep{artetxe-schwenk-2019-massively} and
sentence-BERT~\citep{reimers-2019-sentence-bert} are more suitable for
our pipeline.

We choose the multilingual sentence transformer \citep{reimers-2020-multilingual-sentence-bert} 
version ``\textit{paraphrase-multilingual-mpnet-base-v2}'' as the retriever in our experiments. 
This multilingual retriever is based on XLM-R \citep{conneau-etal-2020-unsupervised} and trained on parallel data from 50+ languages by employing knowledge distillation.
Through the multilingual sentence transformer, sentences are represented by embeddings. We use the sentence embeddings to calculate the cosine similarity between the LRL inputs and HRL sentences and rank the most similar ones for retrieval.
Robustness with respect to other cross-lingual retrievers 
will be discussed in \secref{choiceofre}. 

\paragraph{Multilingual Pretrained Language Model}
In order to solve cloze-style classification tasks, we use the
pretrained multilingual BERT model
``\textit{bert-base-multilingual-cased}'' \citep{devlin-etal-2019-bert}. It
contains 178M parameters and was trained on Wikipedia corpora in 104
languages.  In \secref{choiceofre}, we will also explore XLM-R
\citep{conneau-etal-2020-unsupervised}, another multilingual pretrained
language model.

All the models mentioned above were implemented using the Huggingface Transformers library \citep{wolf2020transformers}.

\section{Results}

\tabref{overview} presents an overview of the results on the three
tasks\footnote{$k=1$ unless otherwise specified.}.  \themethod outperforms the \emph{MAJ}, \emph{Direct} and
\emph{Random} baseline on all three tasks, in both labeled and
unlabeled settings: When retrieving from unlabeled high-resource
language corpora, the performance is improved by \textbf{4.6\%},
\textbf{10.4\%} and \textbf{0.4\%} compared to \emph{Direct} on Amazon
Review, AG News, and XNLI respectively.  When retrieving from labeled
HRL corpora, the performance is improved by
\textbf{15.1\%}, \textbf{31.3\%} and \textbf{2.7\%}.  The
\emph{Finetune} baseline uses the label of retrieved examples for
prompt-based finetuning. Hence it is fair to compare it with
\emph{\themethod} in the labeled setup rather than the unlabeled
one. \emph{\themethod-labeled} outperforms \emph{Finetune} by
\textbf{0.3\%}, \textbf{9.7\%} and \textbf{1.3\%} on the three tasks
respectively.

\renewcommand{\tablename}{Table}
\begin{table}[t]
	\scriptsize
	\centering
	\begin{tabular}{l|c|c|c|c}
		\toprule
		&\textbf{Amazon} & \textbf{AGNews} & \textbf{XNLI} & \textbf{Avg.} \\
		\midrule
		{MAJ} & 50.0& 25.0& 33.3& 36.1\\
		{Random} & 48.2 & 25.6 & 32.4 & 35.4 \\
		{Direct} & 53.8 & 36.3 &33.1&41.1\\
		{Finetune} & 68.6 & 57.9 & 34.5 & 53.7 \\
		\midrule
		{\themethod-unlabeled} & 58.4 & 46.7 & 33.5 &46.2\\
		{\themethod-labeled} & \textbf{68.9} & \textbf{67.6} & \textbf{35.8} &\textbf{57.4}\\
		\bottomrule
	\end{tabular}
	\caption{Overview of results on three classification tasks. 
		The reported numbers are averaged across 10 evaluation LRLs.
	The number of prompts $k$=1 in relevant baselines and our methods for all three tasks.}
	\tablabel{overview}
\end{table}

Although our proposed methods perform better than the baselines on all three tasks, the degree of improvement differs. 
A large improvement is found on AG News, 
the topic categorization task, 
while XNLI only shows a slight improvement. 
An explanation for this could be that the natural language inference task is more difficult than topic categorization, especially in a zero-shot setup. 
Also, prior work has shown that designing cloze-style patterns and searching the answer space for NLI tasks \citep{schick-schutze-2021-exploiting, webson-pavlick-2022-prompt} is difficult.

We also find that \themethod-labeled 
noticeably 
outperforms \themethod-unlabeled,
indicating that the performance of self-prediction is limited by the capabilities of mBERT. More powerful MPLMs and better pattern designs might further improve the performance.

\renewcommand{\tablename}{Table}
\begin{table*}[ht]
\scriptsize
\centering
\begin{tabular}{c|c|c|c|c|c|c|c|c|c|c|c|c|c} 
\toprule
\multicolumn{2}{l|}{} & \underline{\textbf{En}}   & \textbf{Af}   & \textbf{Jv}   & \textbf{Mn}   & \textbf{My}   & \textbf{Sw}   & \textbf{Ta}   & \textbf{Te}   & \textbf{Tl}   & \textbf{Ur}   & \textbf{Uz}   & \textbf{Avg}  \\ 
\midrule
\multicolumn{2}{c|}{MAJ}      & 25.0 & 25.0 & 25.0 & 25.0 & 25.0 & 25.0 & 25.0 & 25.0 & 25.0 & 25.0 & 25.0 & 25.0  \\ 
\hline
\multicolumn{2}{c|}{Direct}   & 52.5 & 41.8 & 27.4 & 42.5 & 32.2 & 31.3 & 31.5 & 33.0 & 31.6 & 46.9 & 44.8 & 36.3  \\ 
\hline
\multirow{6}{*}{UN} & k=1      & 53.7 & 52.8 & 46.2 & 46.5 & 46.1 & 42.8 & 43.3 & 44.3 & 45.0 & 51.0 & 49.7 & 46.7  \\ 
                    & k=3      & 55.8 & 53.6 & 46.2 & 47.1 & 48.2 & 44.9 & 44.5 & 46.3 & 47.1 & 52.6 & 51.0 & 48.1  \\ 
                    & k=5      & 57.1 & 54.4 & 47.0 & 47.0 & 48.0 & 46.6 & 44.8 & 45.8 & 48.5 & 53.1 & 52.3 & 48.7  \\ 
                    & k=10     & 57.5 & 55.3 & 46.3 & 46.4 & 47.6 & 45.6 & 44.1 & 46.7 & 47.7 & 53.0 & 51.4 & 48.4  \\ 
                    & k=20     & 59.7 & 57.2 & 48.1 & 46.7 & 50.0 & 47.9 & 46.0 & \textbf{48.9} & 49.6 & 55.4 & 53.2 & 50.3  \\ 
                    & k=30     & \textbf{60.1} & \textbf{57.4} & \textbf{49.0 }& \textbf{47.4} & \textbf{51.1} &\textbf{ 49.2 }& \textbf{47.1 }& 48.7 & \textbf{50.1} & \textbf{56.5} & \textbf{54.4} & \textbf{51.1 } \\ 
\midrule
\multirow{6}{*}{LB} & k=1      & 74.9 & 75.4 & 68.1 & 63.5 & 68.2 & 64.0 & 62.8 & 65.6 & 64.8 & 72.5 & 71.4 & 67.6  \\ 
                    & k=3      & 77.1 & 77.1 & 69.6 & 65.6 & 71.1 & 67.6 & 65.6 & 68.4 & 65.9 & 74.6 & 74.4 & 70.0  \\ 
                    & k=5      & 78.1 & 78.6 & 69.0 & 64.4 & 72.9 & 68.8 & 65.9 & 69.3 & 66.4 & 75.8 & 75.4 & 70.6  \\ 
                    & k=10     & 78.7 & 79.4 & 70.5 & 67.0 & 72.9 & 68.3 & 66.6 & 70.7 & 67.2 & 76.6 & 75.9 & 71.5  \\ 
                    & k=20     & \textbf{79.0} & 79.7 & 70.7 & 67.5 & 72.5 & \textbf{70.0 }& 67.5 & 70.7 & 68.1 & \textbf{77.4} & 76.3 & 72.0  \\ 
                    & k=30     & 79.0 & \textbf{79.7} &\textbf{ 71.3} & \textbf{67.6} & 72.8 & 69.9 & \textbf{68.1} & \textbf{71.1 }& \textbf{69.4} & 77.2 & \textbf{76.7} & \textbf{72.4}  \\
\bottomrule
\end{tabular}
\caption{Results of topic categorization task on AG News dataset. $k$ is the number of retrieved cross-lingual sample. MAJ is the majority baseline. Avg is the average accuracy across 10 LRLs. \underline{En} is the HRL for retrieval. BoR strategy is adopted.}
\tablabel{details}
\end{table*}

To analyze the performance for every language in detail, 
we present the complete experimental results for the topic categorization task on AG News in \tabref{details}. Here, we use the BoR method to take advantage of multiple retrieved HRL sentences.
As expected, \themethod outperforms the \emph{Direct} baseline on all languages in both labeled and unlabeled settings.

However, it is worth noting that the sensitivity to cross-lingual
retrieval differs from language to language.  For some LRLs, e.g.\ Urdu (Ur) and Uzbek (Uz), \themethod's improvement from cross-lingual
retrieval is smaller. Others gain more, e.g.\ Javanese (Jv). Retrieving more samples increases the performance up to $k$=30 except for Telugu (Te) and Swahili (Sw) where the max is reached for $k$=20.
 
We now turn to the following two questions:
1) How does $k$ affect the performance on other tasks than topic categorization?
2) Which LRLs profit most from our \themethod method and which HRLs are best suited to retrieve prompts?

\section{Analysis}
\subsection{Effect of $k$}

\begin{figure}[ht]
    \centering
    \includegraphics[scale=0.47]{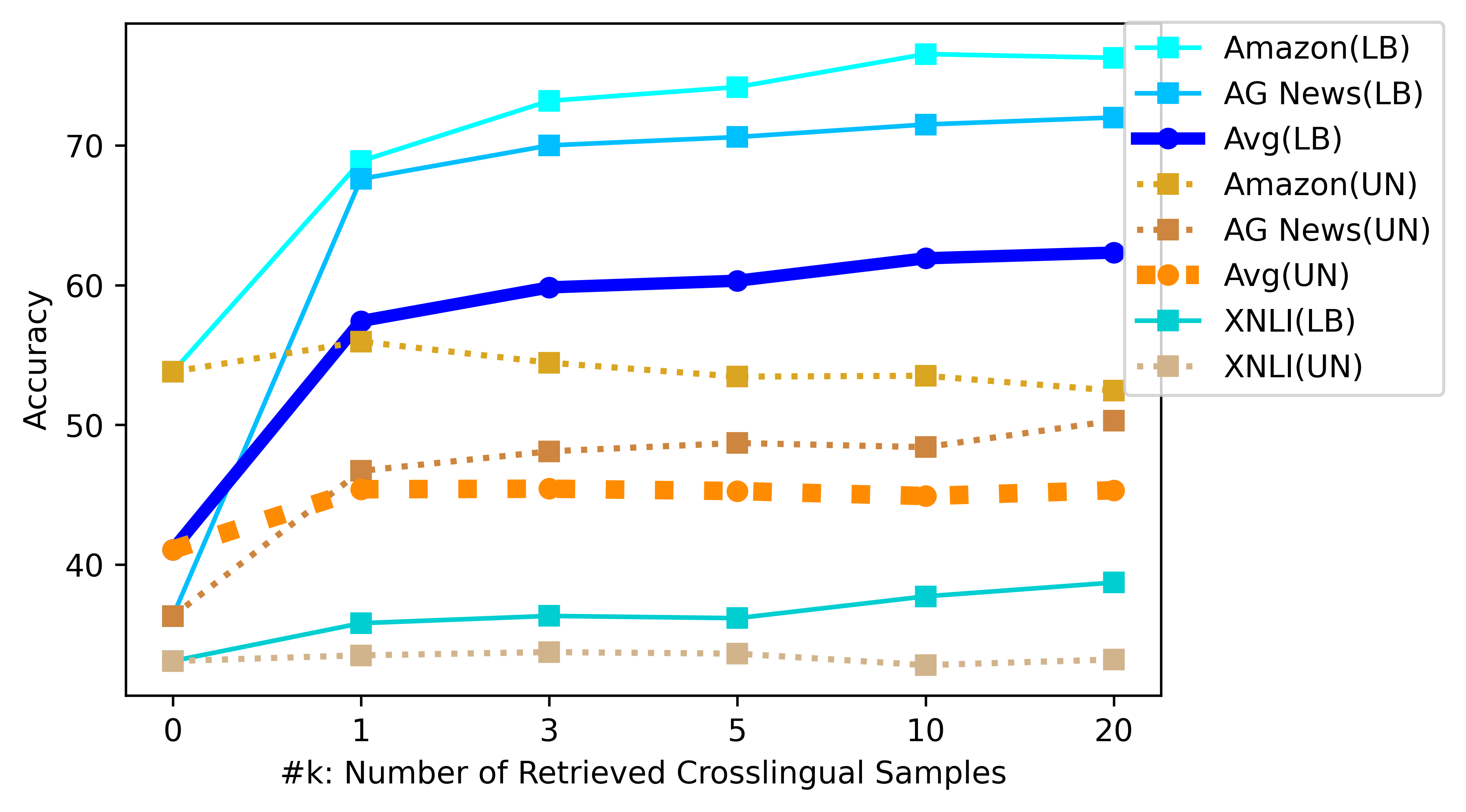}
    \caption{Accuracy on three tasks with different $k$ in the labeled (LB) and unlabeled (UN) setup.}
    \figlabel{analysis-k}
\end{figure}

We investigated how the performance changes as
the number of retrieved HRL samples $k$ increases. 
As shown in \figref{analysis-k}, an abrupt accuracy increase can be seen in both labeled and unlabeled scenarios
by concatenating the most similar cross-lingual sample. 
In labeled scenarios, the performance tends to increase up to $k$=20 and then levels off.  
This can be explained by the fact that later retrieved samples are less similar to the input sample, so their contribution as prompts decreases. 
In unlabeled scenarios, there is no clear improvement beyond k=1 except for
AGNews(UN), where the accuracy increases monotonically except for $k$=10.
The performance of XNLI is less obviously influenced by the value of $k$ than binary sentiment analysis and topic categorization.
We assume that this could be attributed to the difficulty of the inference task. 
Unlike the other two single sentence classification tasks,
XNLI identifies the relationship between a pair of sentences.
Transferring knowledge about sentence relationships is more complicated and requires more samples to learn, in contrast to the other two tasks 
where semantic information from similar cross-lingual sentences can be transferred directly. 







\begin{figure*}[!htb]
\subfigure[Zero-Shot Performance (Unlabeled)]{
\minipage{0.32\textwidth}
  \includegraphics[width=\linewidth]{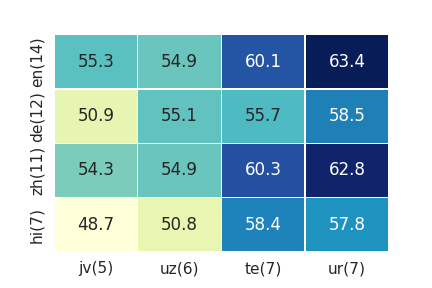}
\endminipage\hfill}
\subfigure[Language Similarity]{
\minipage{0.32\textwidth}
  \includegraphics[width=\linewidth]{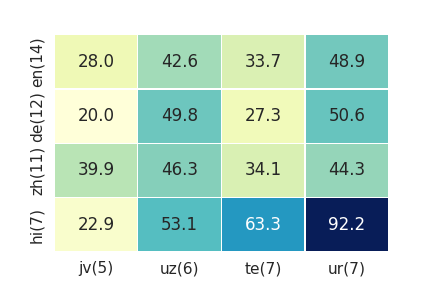}
\endminipage\hfill}
\subfigure[Zero-Shot Performance (labeled)]{
\minipage{0.32\textwidth}%
  \includegraphics[width=\linewidth]{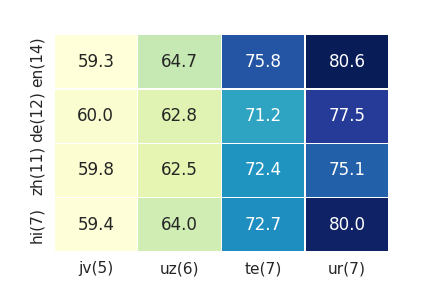}
\endminipage}
	\caption{Visualization of the correlation between zero-shot performance and language similarity, pretraining data size of source and target language. On the X(Y)-axis are target(source) languages with an increasing order of pretraining data size from left(bottom) to right(top). (a) and (c) show the zero-shot performance with \themethod-unlabeled and \themethod-labeled on Amazon review task respectively.  (b) shows the language similarity of each pair.}
    \figlabel{heatmap}
\end{figure*}

\subsection{Effect of Languages}
\seclabel{choiceoflangs}

\citet{lauscher2020zero} pointed out that two linguistic factors exert crucial effects on cross-lingual transfer performance: 
(1) the size of the pretraining corpus for the target language and 
(2) the similarity between the source and target language.
In our study, we also consider a third factor: (3) the size of the pretraining corpus for the source language.
In this section, we conduct a correlation analysis between \themethod's cross-lingual transfer performance and the three language-related factors mentioned above.
To achieve that, we have to measure these factors in a proper way at first.
The size of the pretraining corpus can be easily measured by the $log_2$ value of the Wikipedia size in MB,
as we mentioned in \secref{exp}. 
Thus the remaining problem is how to properly represent language similarity.
\subsubsection{Measurement of Language Similarity}
\citet{malaviya17emnlp} and \citet{littell2017uriel} propose LANG2VEC from linguistic, typological, and phylogenetic perspectives. 
LANG2VEC employs different vectors to represent various types of linguistic features for different languages. Each language is encoded with 5 vectors corresponding to different linguistic features including three typological features (syntax, phonology and phonetic inventory), phylogenetic and geographical features. In typological vectors, each dimension represents a linguistic property. For example, one dimension of the syntax vector represents the word order feature SVO. If a language has a SVO order, then its syntax vector would have the value 1 on this dimension. 
Missing values in the typological vectors could have detrimental
effects. Therefore we replace them with values predicted from the k
most similar typological vectors \citep{malaviya17emnlp}. 
The phylogenetic vector embodies the position of a language in the world language family tree \citep{harald_2015_31872}, while the geographical vector contains the position information of languages w.r.t. their speakers.

Following prior work \citep{Rama2020ProbingMB}, we consider all 5 linguistic features when measuring the language similarity:
syntax (SYN), 
phonology (PHO), 
phonological inventory (INV), 
language family (FAM), and geography (GEO). Given these different types of vectors, we calculate 5 cosine similarities for each pair of source language ($i$) and target language ($j$) and average them to get the final language similarity $sim(i,j)$:

\begin{equation}
    sim(i,j) = \frac{1}{|\mathcal{F}|}\sum_{f\in \mathcal{F}} s(\textbf{v}_f(i), \textbf{v}_f(j))
    \label{sim}
\end{equation}

where $\mathcal{F}$ is the set of features, $\textbf{v}_f(i)$ and $\textbf{v}_f(j)$ stand for the language vectors representing the feature $f$ for $i$ and $j$, and $s(\cdot)$ computes the min-max normalized cosine similarity of the two vectors.
The detailed cosine similarities between English and 10 LRLs evaluated in our experiment are shown in \tabref{langsim} in Appendix \secref{langsim}.

\renewcommand{\tablename}{Table}
\begin{table}[ht]
	\scriptsize
	\centering
	
	\begin{tabular}{l|cc|cc|cc}
		\toprule
		\textbf{Unlabeled} &
		\multicolumn{2}{c}{Sim.} &
		\multicolumn{2}{c}{source size} &
		\multicolumn{2}{c}{target size}
		\\
		\midrule
		&		corr &		p &		corr &		p &		corr &		p \\
		Spearman &		0.28 &		0.05 &		0.20 &		0.16* &		0.31 &		0.03 \\

		Pearson &		0.27 &		0.06* &		0.22 &		0.12* &		0.38 &		6e-03 \\
		\toprule
		\textbf{ labeled } &
		\multicolumn{2}{c}{Sim.} &
		\multicolumn{2}{c}{source size} &
		\multicolumn{2}{c}{target size}
		\\
		\midrule
		&		corr &		p &		corr &		p &		corr &		p \\
		Spearman &		0.42 &		2e-03 &		0.08 &		0.54* &		0.44 &		1e-03 \\

		Pearson &		0.41 &		3e-03 &		-3e-4 &		1.00* &		0.46 &		8e-4 \\
	\bottomrule[1pt]
	\end{tabular}
	\caption{Correlations between Amazon review performance and three features. Sim.: language similarity between an LRL and an HRL; source (target) size: the log of the data size (MB) of source (target). *: insignificant result with a $p$ value larger than 0.05.}
	\tablabel{corr}
\end{table}

\renewcommand{\tablename}{Table}
\begin{table}[t]
	\scriptsize
	\centering
	\begin{tabular}{l|c|c|c|c|c}
	\toprule
	\multicolumn{2}{c}{} & \textbf{Amazon} & \textbf{AGNews} & \textbf{XNLI} & \textbf{Avg.} \\
	\midrule
	\multicolumn{2}{l|}{Direct} & 53.8 & 36.2 & 33.1 & 41.0 \\
	\midrule
	\multirow{4}*{UN} & mBERT+pooling & 53.1 & 36.9 & 33.6 & 41.2\\
	~ & {mBERT+distiluse} & 54.7 & 38.4 & 34.0 & 42.3\\
    ~ & {mBERT+paraphrase} & 59.6 & 46.7 & 33.7 & 46.7\\
	~ & {XLM-R+paraphrase} & \textbf{70.1} & \textbf{57.4} & 34.7 & \textbf{54.1}\\
	~ & {mBERT+LaBSE} & 59.4 & 43.8 & \textbf{35.1} & 46.1\\
	\midrule
	\multirow{4}*{LB} & mBERT+pooling & 53.6 & 58.0 & 33.8 & 48.5\\
	~ & {mBERT+distiluse} & 62.8 & 63.8 & 34.6 & 53.7\\
    ~ & {mBERT+paraphrase} & 72.9 & 67.6 & 36.8 & 59.1\\
	~ & {XLM-R+paraphrase} & \textbf{73.0} & 76.0 & 35.7 & 61.6\\
	~ & {mBERT+LaBSE} & 72.2 & \textbf{80.0} & \textbf{37.5} & \textbf{63.2}\\
	\bottomrule
	\end{tabular}
	\caption{Accuracy with different models used in our approach. pooling: cosine similarity of the last hidden states from the MPLM; distiluse: \textit{distiluse-base-multilingual-cased-v2}, sentence transformer of multilingual distilBERT; paraphrase: \textit{paraphrase-multilingual-mpnet-base-v2}, sentence transformer of XLM-R. UN: unlabeled setup; LB: labeled setup.
	}
	\tablabel{models}
\end{table}

\subsubsection{Correlation Analysis}

We conduct a correlation analysis between cross-lingual performance and the three language factors mentioned above: language similarity between the \textit{source} (retrieved) and \textit{target} (input) language, pretraining data size of the source language and of the target language. We use the log value of Wikipedia size to represent the size of pretraining corpus for target and source languages and $sim(i,j)$ computed by Eq. \eqref{sim} to represent the similarity between the source and target language. Four other HRLs -- Chinese, German, Hindi, Cebuano --  are selected as source languages in addition to English. 
 We measure the cross-lingual performance of \themethod on the Amazon product review task in both the labeled and the unlabeled settings. Full results can be found in Appendix \secref{full_results_corr}.

\tabref{corr} shows the outcome of the correlation analysis. We observe a significant positive correlation between cross-lingual performance and language similarity as well as target language pretraining data size, in both the labeled and the unlabeled setting. The correlation between performance and source language size is not significant.
\figref{heatmap} visualizes the correlations and further clarifies the findings by selecting 4 source languages and 4 target languages and showing the cross-lingual performance and similarity between them. 

\renewcommand{\tablename}{Table}
\begin{table}[ht]
\scriptsize
\centering
\begin{tabular}{l|c|c|c|c|c|c}
\toprule
\multicolumn{2}{c|}{}         & \textbf{Ig}   & \textbf{Sn}   & \textbf{Mt}   & \textbf{Co}   & \textbf{Sm}    \\
\hline
\multicolumn{2}{l|}{Direct}   & 30.3 & 32.1 & 29.8 & 32.6 & 30.4  \\
\hline
\multirow{3}{*}{LB} & k=1    & 56.5 & 59.7 & 63.9 & 75.0 & 52.0  \\
                    & k=3    & 58.1 & 61.4 & 65.2 & 78.2 & 54.1  \\
                    & k=5    & \textbf{58.8} & \textbf{61.6} & \textbf{65.9} & \textbf{79.8} & \textbf{55.4}  \\
\hline
\multirow{3}{*}{UN} & k=1    & 36.6 & 37.3 & 39.1 & 42.6 & 34.4  \\
                    & k=3    & 34.8 & 36.2 & 37.6 & 40.6 & 33.9  \\
                    & k=5    & 34.8 & 35.3 & 37.2 & 40.4 & 34.1  \\
\midrule
\multicolumn{2}{c|}{}         & \textbf{St}   & \textbf{Haw}  & \textbf{Zu}   & \textbf{Ny}   & \textbf{Avg.}  \\
\hline
\multicolumn{2}{l|}{Direct}   & 30.4 & 27.1 & 34.4 & 29.8 & 30.8  \\
\hline
\multirow{3}{*}{LB} & k=1    & 53.5 & 49.9 & 58.0 & 54.9 & 58.1  \\
                    & k=3    & 55.5 & 49.7 & 58.5 & 57.0 & 59.7  \\
                    & k=5    & \textbf{56.8} & \textbf{51.4 }& \textbf{58.8} & \textbf{58.0} & \textbf{60.7}  \\
\hline
\multirow{3}{*}{UN} & k=1    & 36.3 & 31.6 & 35.6 & 35.3 & 36.5  \\
                    & k=3    & 33.7 & 31.0 & 34.3 & 32.9 & 35.0  \\
                    & k=5    & 34.2 & 30.6 & 34.0 & 32.0 & 34.7 \\
\bottomrule
\end{tabular}
\caption{Results of several unseen languages on a topic categorization task (AG News dataset). Ig - Igbo, Sn - Shona, Mt - Maltese, Co - Corsican, Sm - Samoan, St - Sesotho, Haw - Hawaiian, Zu - Zulu, Ny - Chiechewa.}
\tablabel{unseen_langs}
\end{table}

\begin{table*}[!t]
\scriptsize
\centering
\begin{tabular}{c|l|ll|ll|ll|ll|ll} 
\toprule
\multicolumn{2}{c|}{\multirow{2}{*}{}} & \multicolumn{2}{c|}{p1} & \multicolumn{2}{c|}{p2} & \multicolumn{2}{c|}{p3} & \multicolumn{2}{c|}{p4} & \multicolumn{2}{c}{Avg}  \\ 
\cline{3-12}
               \multicolumn{2}{c|}{}  & en & te                 & en & te                 & en & te                 & en & te                 & en & te                   \\ 
\midrule
\multirow{3}{*}{Finetune} & Direct            & \underline{84} & \underline{76}                 & \underline{83} & \underline{70}                 & \underline{86} &  \underline{67}                 &  \underline{85} &  \underline{73}                 &  \underline{85} &  \underline{74}                   \\ 
& \themethod-UN         & 84 -- & 65$\downarrow$                 & 85$\uparrow$ & 62$\downarrow$                 & 83$\downarrow$ & 60$\downarrow$                 & 82$\downarrow$ & 64$\downarrow$                 & 84$\downarrow$ & 67$\downarrow$                   \\ 
& \themethod-LB          & 83$\downarrow$ & 64$\downarrow$                 & 83 -- & 64$\downarrow$                 & 83$\downarrow$ & 64$\downarrow$                 & 82$\downarrow$ & 70$\downarrow$                 & 83$\downarrow$ & 69$\downarrow$                   \\
\midrule

\multirow{3}{*}{w/o Finetune} & Direct            & \underline{54} & \underline{53}                 & \underline{59} & \underline{54}                 & \underline{54} & \underline{50}                 & \underline{53} & \underline{51}                 & \underline{55} & \underline{52}                   \\ 
& \themethod-UN         & 59$\uparrow$ & 55$\uparrow$                 & 55$\downarrow$ & 58$\uparrow$                 & 52$\downarrow$ & 52$\uparrow$                 & 53 -- & 52$\uparrow$                 & 55 -- & 54$\uparrow$                   \\ 
& \themethod-LB          & \textbf{90}$\uparrow$ & \textbf{82}$\uparrow$           & \textbf{90}$\uparrow$ & \textbf{82}$\uparrow$           & \textbf{90}$\uparrow$ & \textbf{82}$\uparrow$    & \textbf{90}$\uparrow$ & \textbf{82}$\uparrow$   & \textbf{90}$\uparrow$ & \textbf{82}$\uparrow$           \\

\bottomrule

\end{tabular}
\caption{Result of English and Telugu on Amazon review task using MPLMs with and without finetuning on English train set. UN: Unlabeled, LB: labeled. $p_i$ represents different prompt patterns.}
\tablabel{ft_result}
\end{table*}

\subsection{Robustness}
\seclabel{choiceofre}

In this section, we test the robustness of the \themethod method w.r.t.\ other cross-lingual retrievers and MPLMs as well as unseen languages.
\subsubsection{Retriever and MPLM}
Apart from the multilingual sentence transformer based on XLM-R (``paraphrase'')  used in our previous experiments,
we explore several other types of cross-lingual retrievers: a ``pooling'' retriever which averages the last hidden states of the MPLM and computes the cosine similarity between these pooled sentence representations; ``distiluse'' retriever, a sentence transformer based on multilingual distilBERT \citep{sanh2019distilbert}; and the ``LaBSE'' retriever \citep{Feng2020LanguageagnosticBS}, a BERT-based model trained for sentence embedding for 109 languages.
As an alternative to mBERT, we also investigate the performance of XLM-R, which has the same architecture as mBERT but is more powerful.
We follow the setup described in \secref{exp}. 

Results are shown in \tabref{models}.
We can find that even the worst combination -- \emph{mBERT+pooling} -- outperforms the \emph{Direct} baseline on average under both labeled and unlabeled settings. 
If the retriever is replaced by a slightly more powerful one, such as the combination \emph{mBERT+distiluse}, higher accuracies in the unlabeled and labeled setting are achieved,
suggesting that our proposed method \themethod is robust w.r.t.\ other cross-lingual retrievers. 
In the result of \emph{XLM-R+paraphrase}, the obviously better performance of XLM-R in the unlabeled setup shows that a stronger MPLM can noticeably improve the self-prediction.
We expect that an even better 
performance could be obtained by applying our proposed \themethod approach to larger and/or more powerful MPLMs such as InfoXLM \citep{chi-etal-2021-infoxlm}. 

\subsubsection{Unseen Languages}

Our previous experiments show that the LRLs pretrained by MPLMs can benefit well from \themethod. However, popular MPLMs are pretrained only on approx. 100 languages, accounting for a tiny part of all languages in the world ($\sim$100/7000). 
We wonder if our proposed method could potentially benefit a wider range of LRLs, so we apply \themethod to several unseen LRLs, i.e. languages not included in the pretrained corpora of the MPLM. 
We conduct experiments on a topic categorization task for nine unseen languages. The results in \tabref{unseen_langs} show that PARC is also effective for unseen LRLs.
It can be observed from the result that \themethod is also effective for unseen LRL languages.

\subsection{Zero-shot Setting}
\seclabel{zeroshot_setting}

Different from the cross-lingual transfer paradigm where a MPLM is first finetuned on annotated training data of one language, and then directly applied to the test data of other languages for inference, our proposed approach is employed in the zero-shot setting for LRLs, i.e., the model parameters are not adjusted by finetuning with HRL data.
\tabref{ft_result} shows results from a preliminary experiment where our PARC
method combined with a finetuned MPLM fails to outperform the Direct
baseline. 
When using finetuned MPLM to evaluate with \themethod, we do not see sufficient performance improvement. However, without finetuning, \themethod performs better in both unlabeled and labeled setup, and \themethod-LB without finetuning also outperforms it with finetuning.

\subsection{Qualitative Analysis}

\renewcommand{\tablename}{Table}
\begin{table}[ht]
\footnotesize
\centering
\begin{tabular}{l}
\textbf{Amazon Review}\\
\toprule
  \textbf{Case \#963} \\
 \textbf{Input:}\\ \includegraphics[scale=0.5]{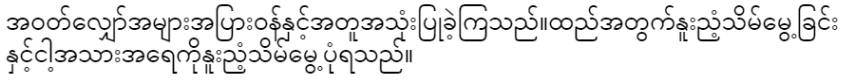}  \\
 (Used with several loads of laundry. Gentle on the fabric \\ and gentle on my skin.) \textcolor{green}{pos}\\
 \textbf{Retrieved}:\\
 \textbf{R1}:  Hard to wash. The fur on top gets all over the sides in \\ the wash. :/ \textcolor{green}{pos}\\
  \textbf{R2}: Very nice and thick high quality towels. \textcolor{green}{pos}\\
  \textbf{R3}: Smelled really bad mold! I had to wash them before \\ use. \textcolor{red}{neg}\\
  \textbf{Predictions:} 
  \textbf{No retrieval - }\textcolor{red}{neg}, \textbf{k=1 - } \textcolor{red}{neg}, \textbf{k=3 - }\textcolor{green}{pos}\\
  \bottomrule
  \end{tabular}
  \caption{A \themethod pipeline example for Amazon review task in the labeled setting.}
  \tablabel{case1}
  \end{table}  

\tabref{case1} shows results of the PARC pipeline for an example from the
Amazon review task. The review in Telugu is positive, but the class
predicted without cross-lingual context is negative. The prediction
stays the same when a single positive English sample is added as
prompt context. When two more English samples are added, the
prediction becomes correct.

 This case indicates that the retrieved cross-lingual samples help the MPLM make a correct decision. Furthermore, more similar HRL samples could rectify the deviation. More cases are shown in \tabref{full_case1} and \tabref{full_case2} in Appendix \secref{case_study}.

\section{Conclusion}
We propose \themethod, a pipeline that augments prompts for zero-shot learning on low resource languages by retrieving semantically similar cross-lingual sentences from HRL corpora. We test \themethod on three classification tasks with parallel test sets across 10 LRLs, and it performs better than the baselines in both unlabeled and labeled settings. Increasing the number of retrieved prompts improves performance at first, but deteriorates it after a certain point. A robustness study shows that \themethod also performs well with other cross-lingual retrievers or MPLMs, suggesting potential applications of \themethod to a wider scope of scenarios.

\section*{Limitations}
The \themethod pipeline proposed in this work is designed to improve the cross-lingual transfer performance for low-resource languages in a zero-shot setting. We tested our method on different LRLs contained in MPLMs and also investigate its effectiveness for several unseen languages. These are not included in pretraining corpora of the MPLM but use a seen script and share some subwords with the seen languages. However, our proposed method is not applicable for unseen languages with new scripts, which restricts its extension towards a wider range of languages. Besides, \themethod is a retrieval-based method. More time and computational resources are required in the cross-lingual retrieval phase. Therefore, it is computationally less efficient to use \themethod for inference. 

\section*{Acknowledgements}
This work was supported by European Research Council (\# 740516), Munich Center for Machine Learning (MCML) and China Scholarship Council (CSC).

\bibliography{anthology,custom, ref}
\bibliographystyle{acl_natbib}
\clearpage

\appendix

\section{Effect of Translations}
\seclabel{translation}
  In our experiment, we use multilingual parallel test sets created by machine translation from English to target low-resource languages. To explore the effect of machine translation-created test sets, we compare the cross-lingual transfer performance on translation test sets and original test sets of XNLI. The original XNLI datasets include two low-resource languages that we used in our experiments, i.e., Swahili (sw) and Urdu (ur). We also measure the translation quality by using the original sets as gold standard. The analysis results (\tabref{translate_analysis}) suggests that machine translated test sets are useful as a proxy for evaluating cross-lingual performance on LRLs.

\renewcommand{\tablename}{Table}
\begin{table}[ht]
\centering
\footnotesize
\begin{tabular}{l|l|r|r} 
\hline
\multicolumn{2}{l|}{Languages}                & \multicolumn{1}{l|}{sw} & \multicolumn{1}{l}{ur}  \\ 
\hline
\multirow{4}{*}{Performance}  & MT Acc.      & 34.00                   & 33.92                    \\ 
                                     & OV Acc.      & 34.07                   & 33.87                    \\ 
                                     & Diff    & 0.07                    & -0.05                    \\ 
                                     & P-Value & 0.85                    & 0.92                     \\ 
\hline
\multirow{3}{*}{Translation Quality} & BLEU    & 56.39                   & 64.96                    \\ 
                                     & chrF    & 49.58                   & 59.89                    \\ 
                                     & Sim.    & 81.82                   & 81.19                    \\
\hline
\end{tabular}
\caption{Comparison of performance on machine translation-created XNLI test sets (MT) and the original version of XNLI test sets (OV) in sw and ur languages. BLEU \& chrF scores and semantic similarities (Sim.) are computed to measure the translation quality of machine translation-created test sets.}
\tablabel{translate_analysis}
\end{table}

		

\section{Language Features}
\seclabel{langsim}
\tabref{langsim} shows the language features of all 10 LRLs evaluated in our experiments. Language similarity refers to the similarity between each LRL and English. SIM score is computed by Eq. \eqref{sim}. WikiSize is the log value of the Wikipedia size in MB.

\begin{table}[t]
\centering
\scriptsize
\begin{tabular}{c|c|c|c|c|c|c|c} 
\toprule
\multirow{2}{*}{\textbf{Lang}} & \multicolumn{6}{c|}{\textbf{Language Similarity}} & \multirow{2}{*}{\begin{tabular}[c]{@{}c@{}}\textbf{Wiki}\\\textbf{Size}\end{tabular}}  \\ 
\cline{2-7}
                               & SYN   & PHO   & INV   & FAM   & GEO   & SIM    &                                                                                        \\ 
\midrule
Af                             & 84.9 & 60.3  & 38.4  & 50.4 & 33.1 & 53.4  & 6                           \\ 
Jv                        & 48.0 & 39.2  & 52.7  & 0.0  & 0.0  & 28.0  & 5                           \\ 
Mn                        & 31.0 & 100.0 & 39.4  & 0.0  & 56.8 & 45.4  & 5                           \\ 
My                        & 17.4 & 80.3  & 100.0 & 0.0  & 37.6 & 47.1 
 & 5                           \\ 
Ta                        & 28.9 & 60.3  & 51.5  & 0.0  & 72.7 & 42.7 & 7                           \\ 
Te                        &  36.0 & 56.2  & 31.3  & 0.0  & 45.2 & 33.7 & 7                           \\ 
Tl                        & 35.0 & 70.5  & 26.7  & 0.0  & 38.8 & 34.2  & 6                           \\ 
Sw                        & 27.0 & 87.0  & 62.1  & 0.0  & 57.2 & 46.6 & 5                           \\ 
Ur                        & 50.2 & 72.0  & 47.1  & 12.6 & 62.5 & 48.9 & 7                           \\ 
Uz                        & 39.8 & 75.6  & 24.1  & 0.0  & 73.7 & 42.6 & 6   \\
\bottomrule
\end{tabular}
\caption{List of language features of the 10 LRLs that we evaluate.}
\tablabel{langsim}
\end{table}

\section{Case Study}
\seclabel{case_study}
\tabref{full_case1} shows two examples from the Amazon Review task. We compare the predictions for three scenarios: no retrieval information (i.e., Direct baseline, see \secref{baseline}), one retrieved sample, and three retrieved samples. Similarly, \tabref{full_case2} shows the same comparison on the AG News task.

\renewcommand{\tablename}{Table}
\begin{table}
\footnotesize
\centering
\begin{tabular}{l}
\textbf{Amazon Review}\\
\toprule
 \textbf{Case 1 \#37} \\
 \textbf{Input:}\\ \includegraphics[scale=0.5]{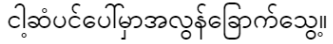}  \\
 (Very dry on my hair.) \textcolor{red}{neg}\\
 \textbf{Retrieved}:\\
 \textbf{R1}: It's a little bit too greasy in my opinion. Doesn't really \\ seem to soak into the hair very well. \textcolor{green}{pos}\\
  \textbf{R2}: The tiniest amount leaves my hair stringy and oily. \textcolor{red}{neg}\\
  \textbf{R3}: could smell this stuff all day but I don’t feel like it \\ moisturizes my skin enough, and my skin isn’t overly dry \\ to begin with. \textcolor{green}{pos}\\
  \textbf{Predictions:} 
  \textbf{No retrieval - }\textcolor{green}{pos}, \textbf{k=1 - } \textcolor{red}{neg}, \textbf{k=3 - }\textcolor{red}{neg}\\
  \toprule
  \textbf{Case 2 \#963} \\
 \textbf{Input:}\\ \includegraphics[scale=0.5]{image/my2.png}  \\
 (Used with several loads of laundry. Gentle on the fabric \\ and gentle on my skin.) \textcolor{green}{pos}\\
 \textbf{Retrieved}:\\
 \textbf{R1}:  Hard to wash. The fur on top gets all over the sides in \\ the wash. :/ \textcolor{green}{pos}\\
  \textbf{R2}: Very nice and thick high quality towels. \textcolor{green}{pos}\\
  \textbf{R3}: Smelled really bad mold! I had to wash them before \\ use. \textcolor{red}{neg}\\
  \textbf{Predictions:} 
  \textbf{No retrieval - }\textcolor{red}{neg}, \textbf{k=1 - } \textcolor{red}{neg}, \textbf{k=3 - }\textcolor{green}{pos}\\
  \hline\hline
  \end{tabular}
  \caption{\themethod examples for Amazon Review task.}
  \tablabel{full_case1}
  \end{table}
  
  \begin{table}[t]
    \centering
    \footnotesize
  \begin{tabular}{l}
  \textbf{AG News}\\
  \toprule
\textbf{Case 1 \#1939} \\
 \textbf{Input:}\\ \includegraphics[scale=0.5]{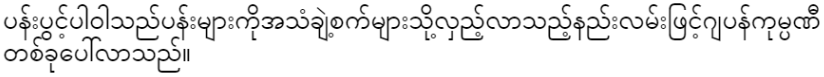}  \\
 (Flower Power A Japanese company has come up with a \\ way to turn flowers into amplifiers. ) \textcolor{blue}{Tech}\\
 \textbf{Retrieved}:\\
 \textbf{R1}: Japanese firms step up spending Japanese firms \\ continue to spend on new equipment and production plants, \\ a survey finds, underlining a continuing recovery in the \\ world's second-largest economy.  \textcolor{blue}{Business}\\
  \textbf{R2}: IBM, Honda deliver in-car speech-recognition \\ navigation system IBM and Honda have jointly developed \\ a hands-free and natural sounding in-vehicle speech-\\ recognition system that will be offered as standard equip-\\ment on the 2005 Acura RL \textcolor{blue}{Tech}\\
  \textbf{R3}: Scientists Make Phone That Turns Into a Sunflower \\ (Reuters) Reuters - Scientists said on Monday they have \\ come up with a cell phone cover that will grow into a \\ sunflower when thrown away. \textcolor{blue}{Tech}\\
  \textbf{Predictions:} 
  \textbf{No retrieval - }\textcolor{red}{World}, \textbf{k=1 - } \textcolor{green}{Tech}, \\ \textbf{k=3 - }\textcolor{green}{Tech}\\
  \toprule
  \textbf{Case 2 \#1302} \\
 \textbf{Input:}\\ \includegraphics[scale=0.5]{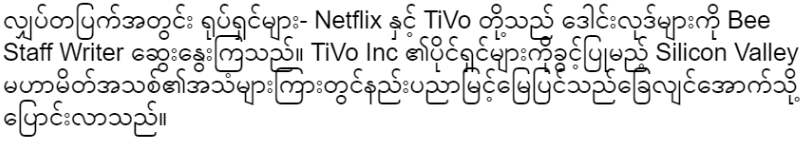}  \\
 (Movies in a Snap: Netflix and TiVo Discuss Downloads \\ Bee Staff Writer. The high-tech terrain is shifting under-\\ foot amid rumblings of a new Silicon Valley alliance \\  that would allow the owners of TiVo Inc. ) \textcolor{blue}{Business}\\
 \textbf{Retrieved}:\\
 \textbf{R1}: NETFLIX, TIVO HOOKUP CLOSE Netflix and \\ TiVo are in late-stage talks on a partnership that would \\ let subscribers use the Internet to download Netflix \\ movies directly into their TiVo box, The Post has \\ learned.  \textcolor{blue}{Business}\\
  \textbf{R2}: TiVo and NetFlix: Picture-Perfect Duo? With TiVo \\ (TIVO) and NetFlix (NFLX ) finally announcing a long-\\ rumored partnership to launch a video-on-demand service \\ sometime next year, investors smiled on the deal that will \\ keep the two popular, but under-fire, innovators ahead of \\ competitors. \textcolor{blue}{Tech}\\
  \textbf{R3}: New Treo and more unveiled at CTIA CTIA stands \\ for the Cellular Telecommunications and Internet \\ Association. Each year they host two shows for the \\ industry. This week is their fall Wireless IT and Enter-\\ tainment expo in San Francisco. \textcolor{blue}{Business}\\
  \textbf{Predictions:} 
  \textbf{No retrieval - }\textcolor{red}{World}, \textbf{k=1 - } \textcolor{green}{Tech}, \\ \textbf{k=3 - }\textcolor{green}{Business}\\
  \hline\hline

\end{tabular}

      \caption{\themethod examples for AG News task}
    \tablabel{full_case2}

\end{table}

\section{Detailed Results}

\subsection{Results for each task}

We show the detailed experimental results for all tasks in \tabref{amazon} (Amazon reviews), \tabref{agnews} (AG News) and \tabref{XNLI} (XNLI), respectively.

\subsection{Detailed data for Correlation Analysis}
\seclabel{full_results_corr}
\tabref{corr_analysis} shows the detailed data used for correlation analysis of language similarity, high- and low-resource language pretraining data size with cross-lingual performance in the unlabeled setting as well as labeled setting.

\subsection{Complete Results for Robustness Analysis}
\tabref{robust_results} shows the results of each language using different combinations of retriever and MPLM for validating the robustness on three tasks.

\renewcommand{\tablename}{Table}

\begin{table*}[t]
\scriptsize
\centering
\begin{tabular}{l|l|ccccc|ccccc|ccccc} 

pattern 0                  & \multicolumn{16}{l}{{[}X] [MASK]}                                                                                     \\ 
pattern 1                  & \multicolumn{16}{l}{It was [MASK]. [X]}                                                                               \\ 
pattern 2                  & \multicolumn{16}{l}{{[}X] All in all, it was [MASK].}                                                                 \\ 
pattern 3                  & \multicolumn{16}{l}{Just [MASK]! [X]}                                                                                \\ 
pattern 4                  & \multicolumn{16}{l}{{[}X] In summary, the product is [MASK].}                                                         \\ 
\toprule
\multicolumn{2}{l|}{\multirow{2}{*}{}}   & \multicolumn{5}{c|}{\underline{\textbf{en}}}          & \multicolumn{5}{c|}{\textbf{af}}          & \multicolumn{5}{c}{\textbf{ur}}           \\ 
\multicolumn{2}{l|}{}                    & p0   & p1   & p2   & p3   & p4   & p0   & p1   & p2   & p3   & p4   & p0   & p1   & p2   & p3   & p4    \\ 
\hline
\multicolumn{2}{l|}{MAJ}                 & 50.0 & 50.0 & 50.0 & 50.0 & 50.0 & 50.0 & 50.0 & 50.0 & 50.0 & 50.0 & 50.0 & 50.0 & 50.0 & 50.0 & 50.0  \\ 

\multicolumn{2}{l|}{Direct}              & 50.5 & 54.3 & 58.9 & 53.7 & 52.6 & 53.3 & 50.7 & 50.4 & 49.8 & 51.5 & 49.9 & 51.7 & 54.6 & 49.9 & 50.3  \\ 
\hline
\multirow{5}{*}{Unlabeled} & k=1          & \textbf{50.9} & \textbf{55.4} & \textbf{59.1} & \textbf{51.9} & \textbf{52.6} & \textbf{51.0} & \textbf{54.9} & \textbf{57.9} & \textbf{52.9} & \textbf{52.8} & \textbf{51.6} & \textbf{56.7} & \textbf{60.0} & \textbf{52.2} & \textbf{52.2}  \\ 
                           & k=3          & 50.7 & 53.7 & 57.7 & 50.8 & 50.4 & 50.4 & 52.5 & 56.2 & 50.7 & 51.0 & 51.3 & 52.9 & 57.1 & 50.8 & 50.9  \\ 
                           & k=5          & 50.8 & 52.2 & 56.0 & 50.3 & 50.9 & 50.8 & 52.2 & 55.0 & 50.2 & 50.6 & 51.2 & 52.5 & 56.4 & 50.3 & 50.7  \\ 
                           & k=10         & 50.7 & 51.9 & 56.0 & 50.0 & 50.6 & 50.7 & 52.0 & 55.8 & 50.2 & 50.7 & 51.4 & 52.4 & 55.5 & 50.0 & 50.3  \\ 
                           & k=20         & 50.5 & 50.8 & 53.6 & 49.9 & 50.1 & 50.5 & 51.1 & 53.5 & 50.0 & 50.2 & 51.1 & 51.2 & 54.0 & 49.8 & 50.0  \\ 
\hline
\multirow{6}{*}{labeled}    & k=1          & \textbf{60.0} & 82.4 & 82.4 & 82.3 & 82.4 & 66.0 & 79.0 & 79.2 & 79.2 & 79.2 & \textbf{57.0} & 80.4 & 80.6 & 80.6 & 80.6  \\ 
                           & k=3          & 58.5 & 86.2 & 86.2 & 86.2 & 86.2 & 65.0 & 80.7 & 81.1 & 81.1 & 81.0 & 56.4 & 83.8 & 84.3 & 84.3 & 84.3  \\ 
                           & k=5          & 57.3 & 87.2 & 87.2 & 87.2 & 87.2 & 65.4 & 82.7 & 82.9 & 82.9 & 82.8 & 56.2 & 84.6 & 85.0 & 85.0 & 85.0  \\ 
                           & k=10         & 57.7 & 88.9 & 88.9 & 88.9 & 88.9 & \textbf{66.5} & 85.2 & 85.4 & 85.4 & 85.4 & 56.6 & 87.0 & 87.3 & 87.3 & 87.3  \\ 
                           & k=20         & 56.4 & \textbf{89.5} & \textbf{89.5} & \textbf{89.5} & \textbf{89.5} & 64.3 & 85.3 & \textbf{85.7} & \textbf{85.7} & \textbf{85.6} & 55.4 & \textbf{87.6} & \textbf{87.9} & \textbf{87.9} & \textbf{88.0}  \\ 
                           & k=30         & 56.3 & 88.9 & 88.9 & 88.9 & 88.9 & 63.6 & 85.4 & 85.6 & 85.6 & 85.6 & 55.7 & 87.4 & 87.6 & 87.6 & 87.6  \\ 
\midrule
\multicolumn{2}{l|}{\multirow{2}{*}{}}   & \multicolumn{5}{c|}{\textbf{sw}}          & \multicolumn{5}{c|}{\textbf{te}}          & \multicolumn{5}{c}{\textbf{ta}}           \\ 

\multicolumn{2}{l|}{}                    & p0   & p1   & p2   & p3   & p4   & p0   & p1   & p2   & p3   & p4   & p0   & p1   & p2   & p3   & p4    \\ 
\hline
\multicolumn{2}{l|}{MAJ}                 & 50.0 & 50.0 & 50.0 & 50.0 & 50.0 & 50.0 & 50.0 & 50.0 & 50.0 & 50.0 & 50.0 & 50.0 & 50.0 & 50.0 & 50.0  \\ 

\multicolumn{2}{l|}{Direct}              & 47.3 & 50.2 & 51.9 & 49.9 & 50.3 & 50.8 & 52.5 & 53.9 & 49.9 & 51.4 & 54.1 & 59.0 & 56.2 & 50.5 & 51.9  \\ 
\hline
\multirow{5}{*}{Unlabeled} & k=1          & \textbf{51.4} & \textbf{50.4} & \textbf{50.5} & \textbf{50.5 }&\textbf{ 50.1} & \textbf{51.6} & \textbf{54.8} & \textbf{57.5} & \textbf{52.3} & \textbf{52.1} & \textbf{57.1} &\textbf{ 55.3} & \textbf{57.2 }& \textbf{52.6} & \textbf{51.6}  \\ 
                           & k=3          & 50.5 & 50.3 & 50.3 & 50.1 & 50.1 & 51.3 & 52.8 & 55.3 & 50.6 & 51.3 & 55.7 & 52.5 & 55.0 & 50.5 & 50.6  \\ 
                           & k=5          & 50.6 & 50.1 & 50.0 & 50.1 & 50.1 & 51.6 & 51.7 & 54.0 & 50.4 & 50.3 & 56.1 & 51.4 & 54.0 & 50.1 & 50.1  \\ 
                           & k=10         & 50.8 & 50.1 & 50.0 & 50.1 & 50.1 & 51.8 & 52.1 & 53.5 & 50.4 & 50.3 & 57.3 & 51.5 & 53.9 & 50.0 & 50.1  \\ 
                           & k=20         & 50.5 & 50.1 & 50.0 & 50.1 & 50.1 & 51.4 & 50.6 & 52.9 & 50.0 & 50.0 & 56.9 & 50.5 & 52.9 & 50.0 & 50.0  \\ 
\hline
\multirow{6}{*}{labeled}    & k=1          & 50.5 & 50.0 & 49.9 & 49.9 & 49.9 & \textbf{58.2} & 75.9 & 75.8 & 75.8 & 75.8 & 68.1 & 75.3 & 75.4 & 75.4 & 75.4  \\ 
                           & k=3          & 51.0 & 54.1 & 54.1 & 54.1 & 54.1 & 58.0 & 78.4 & 78.4 & 78.4 & 78.4 & 70.2 & 79.1 & 79.3 & 79.3 & 79.2  \\ 
                           & k=5          & 50.7 & 54.4 & 54.4 & 54.4 & 54.4 & 56.8 & 79.1 & 79.0 & 79.0 & 79.1 & 70.7 & 80.5 & 80.5 & 80.5 & 80.5  \\ 
                           & k=10         & \textbf{51.3} & \textbf{55.5} &\textbf{ 55.5} & \textbf{55.5} & \textbf{55.5} & 57.2 & 81.3 & 81.6 & 81.6 & 81.6 & \textbf{70.9} & \textbf{83.7} & \textbf{83.9} & \textbf{83.9 }& \textbf{83.9 } \\ 
                           & k=20         & 50.9 & 54.3 & 54.4 & 54.4 & 54.4 & 56.9 & \textbf{82.0} & \textbf{82.1} & \textbf{82.1} & \textbf{82.1 }& 70.8 & 82.8 & 83.1 & 83.1 & 83.1  \\ 
                           & k=30         & 50.7 & 54.3 & 54.3 & 54.3 & 54.3 & 56.8 & 82.0 & 82.0 & 82.0 & 82.0 & 70.5 & 83.3 & 83.5 & 83.4 & 83.4  \\ 
\midrule
\multicolumn{2}{l|}{\multirow{2}{*}{}}   & \multicolumn{5}{c|}{\textbf{mn}}          & \multicolumn{5}{c|}{\textbf{uz}}          & \multicolumn{5}{c}{\textbf{my}}           \\ 

\multicolumn{2}{l|}{}                    & p0   & p1   & p2   & p3   & p4   & p0   & p1   & p2   & p3   & p4   & p0   & p1   & p2   & p3   & p4    \\ 
\hline
\multicolumn{2}{l|}{MAJ}                 & 50.0 & 50.0 & 50.0 & 50.0 & 50.0 & 50.0 & 50.0 & 50.0 & 50.0 & 50.0 & 50.0 & 50.0 & 50.0 & 50.0 & 50.0  \\ 

\multicolumn{2}{l|}{Direct}              & 49.1 & 49.7 & 51.4 & 49.7 & 50.0 & 48.5 & 50.2 & 52.4 & 49.7 & \textbf{51.2} & \textbf{54.4 }& \textbf{56.1} & \textbf{56.1} & 50.5 & \textbf{52.6}  \\ 
\hline
\multirow{5}{*}{Unlabeled} & k=1          & \textbf{51.1} & \textbf{54.7} &\textbf{ 58.6} &\textbf{ 52.6 }& \textbf{52.8 }& 50.4 & \textbf{53.1} & \textbf{53.6} &\textbf{ 51.8} & 50.9 & 53.0 & 53.9 & 56.0 & \textbf{52.3} & 52.0  \\ 
                           & k=3          & 50.2 & 53.2 & 56.4 & 51.0 & 51.1 & 50.5 & 51.9 & 52.1 & 50.2 & 50.3 & 53.0 & 51.5 & 55.0 & 51.2 & 50.7  \\ 
                           & k=5          & 50.2 & 52.0 & 55.3 & 50.4 & 50.5 & 50.5 & 50.3 & 50.7 & 50.0 & 50.2 & 52.9 & 51.1 & 53.6 & 50.5 & 50.3  \\ 
                           & k=10         & 50.4 & 52.2 & 56.3 & \textbf{50.6} & 50.5 & 50.6 & 50.3 & 50.6 & 50.1 & 50.0 & 53.4 & 51.1 & 54.2 & 50.2 & 50.1  \\ 
                           & k=20         & 50.4 & 51.1 & 54.5 & 50.0 & 50.0 & 50.5 & 50.0 & 50.7 & 50.0 & 50.0 & 53.2 & 50.5 & 52.8 & 50.0 & 50.0  \\ 
\hline
\multirow{6}{*}{labeled}    & k=1          & 60.8 & 74.9 & 74.9 & 74.9 & 74.9 & \textbf{56.0} & 65.0 & 64.7 & 64.7 & 64.7 & 65.3 & 73.9 & 73.8 & 73.8 & 73.8  \\ 
                           & k=3          & 60.3 & 79.5 & 79.7 & 79.7 & 79.7 & 55.2 & 65.3 & 65.2 & 65.2 & 65.2 & 66.6 & 77.5 & 77.7 & 77.7 & 77.7  \\ 
                           & k=5          & 59.7 & 80.6 & 80.6 & 80.6 & 80.6 & 55.5 & 66.1 & 66.0 & 66.0 & 65.8 & 65.8 & 78.6 & 78.9 & 78.9 & 78.9  \\ 
                           & k=10         & \textbf{62.2 }& \textbf{83.9} & \textbf{84.3} & \textbf{84.3} & \textbf{84.3 }& 55.9 &\textbf{ 68.1} & \textbf{68.2} & \textbf{68.2 }& \textbf{68.3 }&\textbf{ 67.8} & 80.9 & 81.1 & 81.1 & 81.1  \\ 
                           & k=20         & 60.3 & 82.5 & 83.2 & 83.2 & 83.2 & 53.8 & 67.0 & 67.1 & 67.1 & 67.1 & 67.4 & \textbf{81.8} & \textbf{81.8 }& \textbf{81.8} & \textbf{81.8}  \\ 
                           & k=30         & 59.7 & 83.3 & 83.8 & 83.8 & 83.8 & 54.4 & 67.5 & 67.7 & 67.7 & 67.7 & 67.6 & 81.7 & 81.8 & 81.8 & 81.8  \\ 
\midrule
\multicolumn{2}{l|}{\multirow{2}{*}{}}   & \multicolumn{5}{c|}{\textbf{jv}}          & \multicolumn{5}{c|}{\textbf{tl}}          & \multicolumn{5}{c}{\textbf{Avg.}}         \\ 

\multicolumn{2}{l|}{}                    & p0   & p1   & p2   & p3   & p4   & p0   & p1   & p2   & p3   & p4   & p0   & p1   & p2   & p3   & p4    \\ 
\hline
\multicolumn{2}{l|}{MAJ}                 & 50.0 & 50.0 & 50.0 & 50.0 & 50.0 & 50.0 & 50.0 & 50.0 & 50.0 & 50.0 & 50.0 & 50.0 & 50.0 & 50.0 & 50.0  \\ 

\multicolumn{2}{l|}{Direct}              & \textbf{50.9} & 52.3 & 54.1 & 50.1 & \textbf{52.3} & 49.6 & 50.4 & \textbf{51.9} & 50.0 & \textbf{51.2} & 50.8 & 52.5 & 53.8 & 50.3 & 51.4  \\ 
\hline
\multirow{5}{*}{Unlabeled} & k=1          & 50.6 & \textbf{53.0} & 54.2 & \textbf{50.9} & 50.5 & \textbf{50.4} & \textbf{50.6} & 50.9 & 50.1 & 50.2 & \textbf{51.7} & \textbf{53.9} & \textbf{56.0} & \textbf{51.8} &\textbf{ 51.6}  \\ 
                           & k=3          & 50.2 & 51.7 & \textbf{53.5} & 50.4 & 50.3 & 50.0 & 50.3 & 50.3 & \textbf{50.2} & 50.0 & 51.2 & 52.1 & 54.4 & 50.6 & 50.6  \\ 
                           & k=5          & 50.2 & 50.9 & 52.9 & 50.1 & 50.2 & 50.1 & 50.2 & 50.1 & 50.0 & 50.1 & 51.4 & 51.3 & 53.5 & 50.2 & 50.4  \\ 
                           & k=10         & 50.1 & 50.7 & 52.5 & 49.9 & 50.0 & 50.2 & 50.0 & 50.3 & 50.0 & 50.0 & 51.6 & 51.3 & 53.5 & 50.1 & 50.2  \\ 
                           & k=20         & 50.5 & 50.1 & 51.7 & 50.0 & 50.0 & 50.2 & 50.0 & 50.4 & 50.0 & 50.0 & 51.4 & 50.5 & 52.5 & 50.0 & 50.0  \\ 
\hline
\multirow{6}{*}{labeled}    & k=1          & \textbf{54.1 }& 59.3 & 59.3 & 59.3 & 59.3 & 52.4 & 55.4 & 55.4 & 55.4 & 55.4 & 58.9 & 70.1 & 68.9 & 70.1 & 70.1  \\ 
                           & k=3          & 52.7 & 61.6 & 61.6 & 61.6 & 61.6 & 52.1 & 57.7 & 57.7 & 57.7 & 57.7 & 58.7 & 73.1 & 73.2 & 73.2 & 73.2  \\ 
                           & k=5          & 52.8 & 61.5 & 61.5 & 61.5 & 61.5 & 51.6 & 60.2 & 60.2 & 60.2 & 60.1 & 58.4 & 74.1 & 74.2 & 74.2 & 74.2  \\ 
                           & k=10         & 51.6 & \textbf{62.6} & \textbf{62.6} & \textbf{62.6} & \textbf{62.6} & \textbf{52.4} &\textbf{ 63.2} & \textbf{63.3} & \textbf{63.3 }& \textbf{63.3} &\textbf{ 59.1} & \textbf{76.4} &\textbf{ 76.5 }& \textbf{76.5 }&\textbf{ 76.5 } \\ 
                           & k=20         & 51.6 & 61.5 & 61.5 & 61.5 & 61.5 & 51.5 & 62.8 & 62.9 & 62.9 & 62.9 & 58.1 & 76.1 & 76.3 & 76.3 & 76.3  \\ 
                           & k=30         & 51.6 & 60.9 & 61.0 & 61.0 & 61.0 & 51.5 & 62.3 & 62.4 & 62.4 & 62.4 & 58.0 & 76.1 & 76.2 & 76.2 & 76.2  \\
\bottomrule
\end{tabular}
\caption{Results on Amazon reviews dataset.}
\tablabel{amazon}
\end{table*}

\renewcommand{\tablename}{Table}

\begin{table*}[t]
\scriptsize
\centering
\begin{tabular}{l|l|cccc|cccc|cccc} 
pattern 0                  & \multicolumn{13}{l}{{[}X] [MASK]}  \\     
pattern 1                  & \multicolumn{13}{l}{{[}MASK]: [X]}                                                               \\ 
pattern 2                  & \multicolumn{13}{l}{{[}MASK] News: [X]}                                                          \\ 
pattern 3                  & \multicolumn{13}{l}{{[}X] Category: [MASK]}                                                      \\ 
\toprule
\multicolumn{2}{l|}{\multirow{2}{*}{}}   & \multicolumn{4}{c|}{\underline{\textbf{en}}}   & \multicolumn{4}{c|}{\textbf{af}}   & \multicolumn{4}{c}{\textbf{ur}}    \\ 
\multicolumn{2}{l|}{}                    & p0   & p1   & p2   & p3   & p0   & p1   & p2   & p3   & p0   & p1   & p2   & p3    \\ 
\hline
\multicolumn{2}{l|}{MAJ}                 & 25.0 & 25.0 & 25.0 & 25.0 & 25.0 & 25.0 & 25.0 & 25.0 & 25.0 & 25.0 & 25.0 & 25.0  \\ 
\multicolumn{2}{l|}{Direct}              & 52.5 & 47.8 & \textbf{47.3 }& 53.0 & 41.8 & 41.3 & 40.2 & \textbf{57.8} & 27.4 & 32.4 & 33.0 & \textbf{53.5}  \\ 
\hline
\multirow{5}{*}{Unlabeled} & k=1          & 53.7 & 47.6 & 45.6 & 53.2 & 52.8 & \textbf{46.8} &\textbf{ 46.2} & 53.2 & 46.2 & \textbf{41.8} & \textbf{41.0 }& 49.7  \\ 
                           & k=3          & 55.8 & 47.6 & 43.4 & 54.3 & 53.6 & 46.5 & 44.3 & 54.3 & 46.2 & 40.5 & 38.2 & 49.9  \\ 
                           & k=5          & 57.1 & \textbf{48.3} & 41.7 & \textbf{55.6} & 54.4 & 46.9 & 43.7 & 55.1 & 47.0 & 40.9 & 37.2 & 51.4  \\ 
                           & k=10         & 57.5 & 45.7 & 41.9 & 55.3 & 55.3 & 44.6 & 42.3 & 55.6 & 46.3 & 38.3 & 35.3 & 51.9  \\ 
                           & k=20         &\textbf{ 59.7} & 46.7 & 41.5 & 55.3 & \textbf{57.2} & 45.9 & 42.2 & 56.1 & \textbf{48.1 }& 39.7 & 35.5 & 51.6  \\ 
\hline
\multirow{5}{*}{labeled}   & k=1          & 74.9 & 83.5 & 83.8 & 83.8 & 75.4 & 81.2 & 82.9 & 82.7 & 68.1 & 76.9 & 78.8 & 78.7  \\ 
                           & k=3          & 77.1 & 86.5 & 86.8 & 86.7 & 77.1 & 84.3 & 85.4 & 85.2 & 69.6 & 79.4 & 81.7 & 81.8  \\ 
                           & k=5          & 78.1 & 87.7 & 88.0 & 87.9 & 78.6 & 86.8 & 87.1 & 87.1 & 69.0 & 79.9 & 82.7 & 82.7  \\ 
                           & k=10         & 78.7 & 88.2 & 88.5 & 88.5 & 79.4 & 87.2 & 87.7 & 87.5 & 70.5 & 81.5 & \textbf{83.6 }& \textbf{83.4}  \\ 
                           & k=20         & \textbf{79.0 }&\textbf{ 89.1} & \textbf{89.4} & \textbf{89.4} & \textbf{79.7} & \textbf{87.4} &\textbf{ 87.8 }& \textbf{87.5} & \textbf{70.7 }& \textbf{81.6} & 83.3& 83.2  \\ 
\midrule
\multicolumn{2}{l|}{\multirow{2}{*}{}}   & \multicolumn{4}{c|}{\textbf{sw}}   & \multicolumn{4}{c|}{\textbf{te}}   & \multicolumn{4}{c}{\textbf{ta}}    \\ 

\multicolumn{2}{l|}{}                    & p0   & p1   & p2   & p3   & p0   & p1   & p2   & p3   & p0   & p1   & p2   & p3    \\ 
\hline
\multicolumn{2}{l|}{MAJ}                 & 25.0 & 25.0 & 25.0 & 25.0 & 25.0 & 25.0 & 25.0 & 25.0 & 25.0 & 25.0 & 25.0 & 25.0  \\ 
\multicolumn{2}{l|}{Direct}              & 42.5 & 37.6 & 33.3 & \textbf{56.6} & 32.2 & 37.2 & 32.5 & \textbf{55.4} & 31.3 & 37.2 & 28.6 & 55.1  \\ 
\hline
\multirow{5}{*}{Unlabeled} & k=1          & 46.5 & \textbf{42.1} & \textbf{42.0 }& 46.4 & 46.1 & \textbf{41.5} &\textbf{ 43.3} & 48.6 & 42.8 & \textbf{41.6} &\textbf{ 39.2} & 47.6  \\ 
                           & k=3          & \textbf{47.1} & 41.2 & 39.9 & 47.9 & \textbf{48.2} & 40.0 & 42.4 & 50.3 & 44.9 & 41.0 & 36.9 & 50.1  \\ 
                           & k=5          & 47.0 & 41.5 & 39.3 & 48.6 & 48.0 & 40.4 & 41.0 & 52.4 & 46.6 & 39.8 & 36.0 & 50.9  \\ 
                           & k=10         & 46.4 & 38.5 & 37.0 & 50.0 & 47.6 & 39.0 & 39.3 & 51.8 & 45.6 & 37.8 & 33.9 & 51.5  \\ 
                           & k=20         & 46.7 & 39.1 & 36.9 & 49.9 & 50.0 & 40.1 & 39.7 & 51.6 & \textbf{47.9} & 38.8 & 34.7 & \textbf{52.5 } \\ 
\hline
\multirow{5}{*}{labeled}   & k=1          & 63.5 & 68.4 & 70.3 & 70.3 & 68.2 & 73.9 & 75.0 & 75.0 & 64.0 & 69.7 & 71.5 & 71.5  \\ 
                           & k=3          & 65.6 & 70.8 & 72.3 & 72.4 & 71.1 & 77.6 & 78.2 & 78.2 & 67.6 & 74.4 & 75.7 & 75.7  \\ 
                           & k=5          & 64.4 & 72.2 & 73.5 & 73.4 & \textbf{72.9} & 79.7 & 79.9 & 79.8 & 68.8 & 75.8 & 76.6 & 76.5  \\ 
                           & k=10         & 67.0 & 72.5 & \textbf{74.1 }& \textbf{73.9 }& 72.9 & 79.9 & 80.0 & 80.0 & 68.3 & 76.5 & 77.2 & 77.1  \\ 
                           & k=20         & \textbf{67.5} & \textbf{72.7} & 73.6 & 73.6 & 72.5 & \textbf{80.2} & \textbf{80.6 }& \textbf{80.6 }&\textbf{ 70.0} & \textbf{77.5} & \textbf{78.1} & \textbf{78.2}  \\ 
\midrule
\multicolumn{2}{l|}{\multirow{2}{*}{}}   & \multicolumn{4}{c|}{\textbf{mn}}   & \multicolumn{4}{c|}{\textbf{uz}}   & \multicolumn{4}{c}{\textbf{my}}    \\ 

\multicolumn{2}{l|}{}                    & p0   & p1   & p2   & p3   & p0   & p1   & p2   & p3   & p0   & p1   & p2   & p3    \\ 
\hline
\multicolumn{2}{l|}{MAJ}                 & 25.0 & 25.0 & 25.0 & 25.0 & 25.0 & 25.0 & 25.0 & 25.0 & 25.0 & 25.0 & 25.0 & 25.0  \\ 
\multicolumn{2}{l|}{Direct}              & 31.5 & 30.9 & 32.0 & 47.3 & 33.0 & 37.5 & 33.8 & 50.7 & 31.6 & 37.4 & 33.7 & 51.9  \\ 
\hline
\multirow{5}{*}{Unlabeled} & k=1          & 43.3 & \textbf{42.5} & \textbf{41.5} & 48.2 & 44.3 &\textbf{ 44.4} & \textbf{42.3} & 49.0 & 45.0 & 43.9 &\textbf{ 43.6} & 50.0  \\ 
                           & k=3          & 44.5 & 41.2 & 40.5 & 51.1 & 46.3 & 42.2 & 40.7 & 50.9 & 47.1 & \textbf{44.5} & 41.7 & 53.7  \\ 
                           & k=5          & 44.8 & 41.5 & 39.6 & 51.8 & 45.8 & 41.7 & 39.2 & 52.3 & 48.5 & 43.8 & 41.4 & 54.2  \\ 
                           & k=10         & 44.1 & 39.7 & 38.0 & \textbf{53.3} & 46.7 & 39.7 & 37.9 & \textbf{53.4} & 47.7 & 41.4 & 40.0 & \textbf{54.4}  \\ 
                           & k=20         & \textbf{46.0} & 39.7 & 37.9 & 52.8 &\textbf{ 48.9} & 41.2 & 36.9 & 53.1 & \textbf{49.6} & 42.2 & 40.3 & 53.6  \\ 
\hline
\multirow{5}{*}{labeled}   & k=1          & 62.8 & 70.9 & 72.7 & 72.8 & 65.6 & 71.5 & 73.2 & 73.3 & 64.8 & 76.2 & 77.4 & 77.2  \\ 
                           & k=3          & 65.6 & 75.4 & 77.3 & 77.2 & 68.4 & 73.6 & 75.7 & 75.7 & 65.9 & 79.5 & 80.1 & 79.8  \\ 
                           & k=5          & 65.9 & 75.8 & 78.0 & 77.9 & 69.3 & 76.1 & 77.9 & 77.8 & 66.4 & 81.4 & 82.5 & 81.8  \\ 
                           & k=10         & 66.6 & 77.0 & \textbf{78.7 }& \textbf{78.6} & 70.7 & 76.4 & 78.3 & 78.2 & 67.2 & 82.4 & 82.9 & 82.3  \\ 
                           & k=20         & \textbf{67.5} & \textbf{77.4} & 78.2 & 78.0 & \textbf{70.7} & \textbf{77.3} & \textbf{78.8} & \textbf{78.7 }& \textbf{68.1} &\textbf{ 83.1} & \textbf{83.6 }& \textbf{83.3}  \\ 
\midrule
\multicolumn{2}{l|}{\multirow{2}{*}{}}   & \multicolumn{4}{c|}{\textbf{jv}}   & \multicolumn{4}{c|}{\textbf{tl}}   & \multicolumn{4}{c}{\textbf{Avg}}   \\ 

\multicolumn{2}{l|}{}                    & p0   & p1   & p2   & p3   & p0   & p1   & p2   & p3   & p0   & p1   & p2   & p3    \\ 
\hline
\multicolumn{2}{l|}{MAJ}                 & 25.0 & 25.0 & 25.0 & 25.0 & 25.0 & 25.0 & 25.0 & 25.0 & 25.0 & 25.0 & 25.0 & 25.0  \\ 
\multicolumn{2}{l|}{Direct}              & 46.9 & 39.3 & 38.0 & \textbf{59.3} & 44.8 & 44.4 & 42.6 & \textbf{60.4} & 37.8 & 38.4 & 36.2 & 50.9  \\ 
\hline
\multirow{5}{*}{Unlabeled} & k=1          & 51.0 & \textbf{45.5} & \textbf{45.4} & 51.6 & 49.7 & \textbf{45.8 }& \textbf{43.7} & 52.2 & 47.4 & \textbf{44.2} & \textbf{43.5} & 48.9  \\ 
                           & k=3          & 52.6 & 44.6 & 42.0 & 53.5 & 51.0 & 45.3 & 42.7 & 54.0 & 48.8 & 43.6 & 41.9 & 50.3  \\ 
                           & k=5          & 53.1 & 44.5 & 41.3 & 53.6 & 52.3 & 45.2 & 41.8 & 54.2 & 49.5 & 43.7 & 41.2 & 51.0  \\ 
                           & k=10         & 53.0 & 42.4 & 39.9 & 54.0 & 51.4 & 44.0 & 39.8 & 54.9 & 49.2 & 41.7 & 39.7 & 51.2  \\ 
                           & k=20         & \textbf{55.4 }& 42.8 & 40.1 & 54.2 &\textbf{ 53.2} & 44.4 & 38.9 & 55.3 & \textbf{51.1} & 42.6 & 39.9 & \textbf{51.4}  \\ 
\hline
\multirow{5}{*}{labeled}   & k=1          & 72.5 & 77.8 & 79.1 & 79.1 & 71.4 & 76.6 & 78.9 & 79.0 & 68.3 & 74.6 & 75.9 & 75.9  \\ 
                           & k=3          & 74.6 & 80.5 & 82.3 & 82.3 & 74.4 & 80.7 & 82.1 & 82.2 & 70.6 & 77.8 & 78.9 & 78.9  \\ 
                           & k=5          & 75.8 & 81.3 & 82.8 & 82.8 & 75.4 & 81.2 & 83.4 & 83.5 & 71.3 & 79.1 & 80.2 & 80.1  \\ 
                           & k=10         & 76.6 & 82.0 & 84.0 & 84.2 & 75.9 & 82.4 & \textbf{84.5} & \textbf{84.6} & 72.1 & 79.8 & 80.9 & 80.8  \\ 
                           & k=20         & \textbf{77.4 }& \textbf{82.8} & \textbf{84.6} & \textbf{84.8 }& \textbf{76.3 }& \textbf{82.8} & 84.0 & 84.0 & \textbf{72.6 }&\textbf{ 80.4} & \textbf{81.1} & \textbf{81.1}  \\
\bottomrule
\end{tabular}
\caption{Results on AG News dataset.}
\tablabel{agnews}
\end{table*}


\begin{table*}
\scriptsize
\centering
\begin{tabular}{l|l|ccc|ccc|ccc|ccc} 
pattern 0                  & \multicolumn{13}{l}{{[}$X_1$] [MASK] [$X_2$]}                                                                               \\ 
pattern 1                  & \multicolumn{13}{l}{{[}$X_1$]? [MASK], [$X_2$] (Yes - No)}                                                                  \\ 
pattern 2                  & \multicolumn{13}{l}{{[}$X_1$]? [MASK], [$X_2$] (Right - Wrong)}                                                             \\ 
\toprule
\multicolumn{2}{l|}{\multirow{2}{*}{}}   & \multicolumn{3}{c|}{\underline{\textbf{en}}} & \multicolumn{3}{c|}{\textbf{af}} & \multicolumn{3}{c|}{\textbf{ur}} & \multicolumn{3}{c}{\textbf{sw}}   \\ 
\multicolumn{2}{l|}{}                    & p0   & p1   & p2        & p0   & p1   & p2        & p0   & p1   & p2        & p0   & p1   & p2          \\ 
\hline
\multicolumn{2}{l|}{MAJ}                 & 33.3 & 33.3 & 33.3      & 33.3 & 33.3 & 33.3      & 33.3 & 33.3 & 33.3      & 33.3 & 33.3 & 33.3        \\ 
\multicolumn{2}{l|}{Direct}            & 33.3 & \textbf{34.2} & 34.3      & 33.2 & 33.0 & 33.4      & \textbf{33.6} & 34.0 & 33.2      & 33.2 & 32.2 & 33.1        \\ 
\hline
\multirow{5}{*}{Unlabeled} & k=1          & \textbf{34.1} & 33.7 & \textbf{34.5}      &\textbf{ 34.0} & 3\textbf{4.1} & 33.7      & 32.4 & \textbf{35.3 }& 32.7      & 33.5 &\textbf{ 33.7} & \textbf{33.7 }       \\ 
                           & k=3          & 33.7 & \textbf{34.1} & 34.3      & 33.0 & 32.9 & 34.1      & 33.3 & 34.0 & \textbf{33.9}      & \textbf{33.6} & 33.0 & 33.5        \\ 
                           & k=5          & 31.9 & 33.7 & 34.3      & 32.5 & 32.8 & 33.9      & 31.2 & 34.1 & 33.6      & 33.2 & 32.7 & 32.9        \\ 
                           & k=10         & 31.9 & 33.6 & 33.3      & 31.9 & 33.3 & 32.6      & 32.2 & 34.2 & 33.2      & 33.0 & 32.7 & 32.5        \\ 
                           & k=20         & 32.0 & 34.4 & 33.3      & 31.6 & 33.6 & 34.1      & 31.6 & 34.4 & 33.9      & 33.1 & 33.1 & 32.0        \\ 
\hline
\multirow{5}{*}{labeled}   & k=1          & 38.9 & 39.1 & 38.8      & 38.7 & 38.9 & 38.1      & 37.0 & 37.4 & 36.7      & 33.3 & 33.4 & 33.4        \\ 
                           & k=3          & 39.2 & 39.1 & 38.6      & 37.9 & 37.9 & 37.4      & 37.0 & 37.8 & 36.8      & 33.7 & 33.5 & 33.7        \\ 
                           & k=5          & 40.0 & 39.8 & 39.5      & 38.0 & 38.0 & 37.1      & 40.2 & 40.6 & 39.8      & 32.7 & 32.5 & 32.6        \\ 
                           & k=10         & 41.5 & 41.6 & 40.9      & 41.1 & 41.1 & 40.5      & 42.0 & 42.4 & 41.0      & 33.7 & 33.7 & 34.1        \\ 
                           & k=20         & \textbf{44.5} & \textbf{44.1 }&\textbf{ 43.5}      & \textbf{42.3} & \textbf{43.0} & \textbf{41.3 }     & \textbf{42.4} &\textbf{ 43.4} & \textbf{42.2 }     & \textbf{35.9 }& \textbf{35.7 }& \textbf{35.9 }       \\ 
\midrule
\multicolumn{2}{l|}{\multirow{2}{*}{}}   & \multicolumn{3}{c|}{\textbf{te}} & \multicolumn{3}{c|}{\textbf{ta}} & \multicolumn{3}{c|}{\textbf{mn}} & \multicolumn{3}{c}{\textbf{uz}}   \\ 
\multicolumn{2}{l|}{}                    & p0   & p1   & p2        & p0   & p1   & p2        & p0   & p1   & p2        & p0   & p1   & p2          \\ 
\hline
\multicolumn{2}{l|}{MAJ}                 & 33.3 & 33.3 & 33.3      & 33.3 & 33.3 & 33.3      & 33.3 & 33.3 & 33.3      & 33.3 & 33.3 & 33.3        \\ 
\multicolumn{2}{l|}{Direct}            & 31.9 & 33.0 & 33.2      & 32.4 & 34.1 & 32.9      & \textbf{33.0 }& 32.7 & 32.6      &\textbf{ 33.3} & 33.3 & 32.9        \\ 
\hline
\multirow{5}{*}{Unlabeled} & k=1          & \textbf{34.1} & 34.1 & \textbf{34.1 }     & \textbf{34.5} & 34.3 & 33.3      & 32.8 & 33.6 & \textbf{34.7}      & 33.2 & 33.9 & 32.8        \\ 
                           & k=3          & 32.8 & 34.9 & 33.4      & 33.7 & 34.7 & \textbf{34.2}      & 32.2 & \textbf{34.5} & 33.7      & 32.3 & 34.5 & 33.4        \\ 
                           & k=5          & 32.9 & \textbf{35.1} & 33.8      & 32.9 & 34.3 & 33.9      & 31.9 & 33.9 & 34.1      & 33.1 & \textbf{34.5} & \textbf{33.9  }      \\ 
                           & k=10         & 32.0 & 34.1 & 32.7      & 32.3 & 34.7 & 32.5      & 30.8 & 34.1 & 32.5      & 32.8 & 33.9 & 32.6        \\ 
                           & k=20         & 31.5 & 34.6 & 32.7      & 32.5 & \textbf{34.8} & 32.9      & 32.0 & 34.1 & 33.4      & 32.6 & 33.5 & 32.6        \\ 
\hline
\multirow{5}{*}{labeled}   & k=1          & 37.8 & 38.1 & 37.7      & 37.7 & 38.0 & 37.0      & 36.5 & 36.5 & 36.5      & 35.5 & 34.8 & 35.0        \\ 
                           & k=3          & 38.9 & 39.5 & 38.4      & 38.7 & 39.4 & 37.5      & 39.1 & 39.1 & 38.9      & 35.1 & 34.7 & 34.7        \\ 
                           & k=5          & 37.5 & 37.1 & 35.9      & 38.3 & 38.7 & 36.3      & 37.1 & 36.9 & 36.9      & 36.0 & 35.9 & 35.9        \\ 
                           & k=10         & 39.2 & 39.5 & 37.9      & 41.1 & 40.8 & 38.0      & 39.5 & 39.3 & 39.3      & 38.3 & 37.9 & 37.8        \\ 
                           & k=20         & \textbf{41.2 }& \textbf{41.5 }& \textbf{39.3}      &\textbf{ 42.7} & \textbf{43.1 }& \textbf{39.7  }    & \textbf{40.3} & \textbf{40.2} &\textbf{ 40.0 }     & \textbf{40.0} & \textbf{39.9} &\textbf{ 39.6  }      \\ 
\midrule
\multicolumn{2}{l|}{\multirow{2}{*}{}}   & \multicolumn{3}{c|}{\textbf{my}} & \multicolumn{3}{c|}{\textbf{jv}} & \multicolumn{3}{c}{\textbf{tl}} & \multicolumn{3}{c}{\textbf{Avg}}  \\ 
\multicolumn{2}{l|}{}                    & p0   & p1   & p2        & p0   & p1   & p2        & p0   & p1   & p2        & p0   & p1   & p2          \\ 
\hline
\multicolumn{2}{l|}{MAJ}                 & 33.3 & 33.3 & 33.3      & 33.3 & 33.3 & 33.3      & 33.3 & 33.3 & 33.3      & 33.3 & 33.3 & 33.3        \\ 
\multicolumn{2}{l|}{Direct}            & \textbf{33.7 }& 33.6 & 33.7      & \textbf{33.3} & 33.3 & 33.6      & 33.3 & 33.5 & 32.3      & 33.1 & 33.3 & 33.1        \\ 
\hline
\multirow{5}{*}{Unlabeled} & k=1          & 33.3 & 33.5 & \textbf{33.8 }     & 32.4 & 32.0 & 33.3      & 33.8 & 32.7 & 32.8      & \textbf{33.4 }& 33.7 & 33.5        \\ 
                           & k=3          & 32.6 & 33.9 & 33.7      & 32.1 & 31.4 & 34.2      & 33.7 & \textbf{33.9} & \textbf{33.3}      & 32.9 & \textbf{33.7} &\textbf{ 33.7}        \\ 
                           & k=5          & 32.5 & \textbf{34.3} & 33.6      & 32.4 & 31.6 & 34.3      & \textbf{34.1} & 33.5 & 32.1      & 32.7 & 33.6 & 33.6        \\ 
                           & k=10         & 30.5 & 33.9 & 33.3      & 32.1 & 32.6 & 33.5      & 33.2 & 33.1 & 32.6      & 32.1 & 33.5 & 32.8        \\ 
                           & k=20         & 30.9 & 33.5 & 32.7      & 30.8 & \textbf{33.6} & \textbf{34.7 }     & 32.9 & 32.5 & 33.1      & 32.0 & 33.6 & 33.2        \\ 
\hline
\multirow{5}{*}{labeled}   & k=1          & 36.8 & 36.7 & 36.1      & 34.2 & 33.5 & 33.3      & 34.7 & 34.4 & 34.3      & 36.2 & 36.2 & 35.8        \\ 
                           & k=3          & 36.7 & 36.9 & 36.2      & 34.6 & 33.9 & 33.9      & 35.7 & 35.7 & 35.7      & 36.7 & 36.8 & 36.3        \\ 
                           & k=5          & 37.7 & 37.7 & 37.3      & \textbf{35.2} & \textbf{34.8} & \textbf{34.6}      & 35.7 & 35.7 & 35.3      & 36.9 & 36.8 & 36.2        \\ 
                           & k=10         & 39.5 & 39.3 & 38.1      & 34.7 & 34.4 & 33.6      & 37.2 & 36.9 & 36.9      & 38.6 & 38.5 & 37.7        \\ 
                           & k=20         & \textbf{41.7} & \textbf{41.3} & \textbf{39.6}      & 32.8 & 32.8 & 32.4      & \textbf{37.4} & \textbf{37.0} & \textbf{37.0 }     & \textbf{39.7} & \textbf{39.8} & \textbf{38.7 }       \\
\bottomrule
\end{tabular}
\caption{Results on XNLI dataset.}
\tablabel{XNLI}
\end{table*}

\renewcommand{\tablename}{Table}
\begin{table*}
\scriptsize
\centering
\begin{tabular}{c|l|l|c|c|c|c|c|c|c|c|c|c|c|c} 
\toprule
\multicolumn{3}{c|}{~}                               & \underline{\textbf{En}}   & \textbf{Af}   & \textbf{Jv }  & \textbf{Mn}   & \textbf{My}   & \textbf{Sw}   & \textbf{Ta}   & \textbf{Te}   & \textbf{Tl}   & \textbf{Ur }  & \textbf{Uz}   & \textbf{Avg }  \\ 
\hline
\multicolumn{3}{c|}{MAJ}                             & 25.0 & 25.0 & 25.0 & 25.0 & 25.0 & 25.0 & 25.0 & 25.0 & 25.0 & 25.0 & 25.0 & 25.0  \\ 
\hline
\multicolumn{3}{c|}{Direct}                          & 52.5 & 41.8 & 27.4 & 42.5 & 32.2 & 31.3 & 31.5 & 33.0 & 31.6 & 46.9 & 44.8 & 36.3  \\ 
\hline
\multirow{11}{*}{UN} & \multicolumn{2}{l|}{k=1}      & 53.7 & 52.8 & 46.2 & 46.5 & 46.1 & 42.8 & 43.3 & 44.3 & 45.0 & 51.0 & 49.7 & 46.7  \\
\cline{2-15}
                     & \multirow{2}{*}{k=3}  & BoR  & 55.8 & 53.6 & 46.2 & 47.1 & 48.2 & 44.9 & 44.5 & 46.3 & 47.1 & 52.6 & 51.0 & 48.1  \\
                     &                       & CONC & 53.5 & 52.4 & 45.9 & 44.9 & 44.8 & 42.9 & 41.7 & 46.6 & 46.0 & 52.0 & 51.6 & 46.9  \\
\cline{2-15}
                     & \multirow{2}{*}{k=5}  & BoR  & 57.1 & 54.4 & 47.0 & 47.0 & 48.0 & 46.6 & 44.8 & 45.8 & 48.5 & 53.1 & 52.3 & 48.7  \\
                     &                       & CONC & 53.5 & 48.0 & 38.2 & 41.3 & 36.3 & 36.9 & 39.5 & 41.4 & 42.9 & 50.5 & 49.6 & 42.4  \\
\cline{2-15}
                     & \multirow{2}{*}{k=10} & BoR  & 57.5 & 55.3 & 46.3 & 46.4 & 47.6 & 45.6 & 44.1 & 46.7 & 47.7 & 53.0 & 51.4 & 48.4  \\
                     &                       & CONC & 46.4 & 41.1 & 36.2 & 38.3 & 36.6 & 34.9 & 34.6 & 35.8 & 40.7 & 46.3 & 45.0 & 38.9  \\
\cline{2-15}
                     & \multirow{2}{*}{k=20} & BoR  & 59.7 & 57.2 & 48.1 & 46.7 & 50.0 & 47.9 & 46.0 & 48.9 & 49.6 & 55.4 & 53.2 & 50.3  \\
                     &                       & CONC & 50.0 & 48.4 & 42.3 & 41.4 & 43.3 & 43.1 & 39.3 & 44.3 & 48.1 & 47.9 & 48.4 & 44.6  \\
\cline{2-15}
                     & \multirow{2}{*}{k=30} & BoR  & 60.1 & 57.4 & 49.0 & 47.4 & 51.1 & 49.2 & 47.1 & 48.7 & 50.1 & 56.5 & 54.4 & 51.1  \\
                     &                       & CONC & 50.7 & 47.6 & 43.9 & 38.2 & 42.9 & 42.5 & 41.8 & 44.5 & 47.7 & 47.1 & 47.3 & 44.3  \\
\midrule
\multirow{11}{*}{LB} & \multicolumn{2}{l|}{k=1}      & 74.9 & 75.4 & 68.1 & 63.5 & 68.2 & 64.0 & 62.8 & 65.6 & 64.8 & 72.5 & 71.4 & 67.6  \\
\cline{2-15}
                     & \multirow{2}{*}{k=3}  & BoR  & 77.1 & 77.1 & 69.6 & 65.6 & 71.1 & 67.6 & 65.6 & 68.4 & 65.9 & 74.6 & 74.4 & 70.0  \\
                     &                       & CONC & 75.6 & 74.8 & 67.3 & 63.1 & 60.3 & 59.0 & 60.5 & 67.1 & 65.9 & 73.3 & 72.4 & 66.4  \\
\cline{2-15}
                     & \multirow{2}{*}{k=5}  & BoR  & 78.1 & 78.6 & 69.0 & 64.4 & 72.9 & 68.8 & 65.9 & 69.3 & 66.4 & 75.8 & 75.4 & 70.6  \\
                     &                       & CONC & 74.6 & 66.5 & 48.2 & 53.9 & 44.9 & 45.4 & 52.1 & 59.5 & 56.0 & 70.9 & 63.6 & 56.1  \\
\cline{2-15}
                     & \multirow{2}{*}{k=10} & BoR  & 78.7 & 79.4 & 70.5 & 67.0 & 72.9 & 68.3 & 66.6 & 70.7 & 67.2 & 76.6 & 75.9 & 71.5  \\
                     &                       & CONC & 61.2 & 52.7 & 43.2 & 48.0 & 44.5 & 42.5 & 41.3 & 45.0 & 50.1 & 62.3 & 56.7 & 48.6  \\
\cline{2-15}
                     & \multirow{2}{*}{k=20} & BoR  & 79.0 & 79.7 & 70.7 & 67.5 & 72.5 & 70.0 & 67.5 & 70.7 & 68.1 & 77.4 & 76.3 & 72.0  \\
                     &                       & CONC & 67.4 & 65.1 & 55.8 & 55.6 & 57.6 & 58.3 & 51.2 & 61.0 & 62.8 & 66.4 & 66.0 & 60.0  \\
\cline{2-15}
                     & \multirow{2}{*}{k=30} & BoR  & 79.0 & 79.7 & 71.3 & 67.6 & 72.8 & 69.9 & 68.1 & 71.1 & 69.4 & 77.2 & 76.7 & 72.4  \\
                     &                       & CONC & 72.8 & 71.1 & 62.1 & 57.0 & 61.6 & 60.4 & 57.9 & 67.9 & 64.6 & 71.6 & 69.3 & 64.3 \\
\bottomrule                     
\end{tabular}
\caption{Results of topic categorization task on AG News Dataset. $k$ is the number of retrieved cross-lingual sample.
MAJ is the majority baseline. Avg is the average accuracy across 10 LRLs. En is the HRL for retrieval. BoR refers to the \textit{Bag of Retrieval} strategy, CONC refers to the \textit{Concatenation} strategy.}
\tablabel{agnews_conc}
\end{table*}

\begin{table*}
\scriptsize
\centering
\begin{tabular}{l|c|c|ccccc|c|c|c} 
\toprule
\multirow{2}{*}{} & \multicolumn{2}{c|}{\textbf{Performance}} & \multicolumn{6}{c|}{\textbf{Language Similarity}}    & \multicolumn{2}{c}{\textbf{WikiSize}}  \\ 
\cline{2-11}
                  &              Unlabeled & labeled                & SYN  & PHO   & INV  & FAM   & GEO  & SIM  & source & target                       \\ 
\midrule
en-af  & 79.2 & 62.0 & 84.9  & 60.3  & 38.4  & 50.4  & 33.1  & 53.4 & 14 & 6  \\
en-ur  & 80.6 & 63.4 & 50.2  & 72.0  & 47.1  & 12.6  & 62.5  & 48.9 & 14 & 7  \\
en-sw  & 49.9 & 51.0 & 27.0  & 87.0  & 62.1  & 0.0   & 57.2  & 46.6 & 14 & 5  \\
en-te  & 75.8 & 60.1 & 36.0  & 56.2  & 31.3  & 0.0   & 45.2  & 33.7 & 14 & 7  \\
en-ta  & 75.4 & 60.2 & 28.9  & 60.3  & 51.5  & 0.0   & 72.7  & 42.7 & 14 & 7  \\
en-mn  & 74.9 & 62.9 & 31.0  & 100.0 & 39.4  & 0.0   & 56.8  & 45.4 & 14 & 5  \\
en-uz  & 64.7 & 54.9 & 39.8  & 75.6  & 24.1  & 0.0   & 73.7  & 42.6 & 14 & 6  \\
en-my  & 73.8 & 60.3 & 17.4  & 80.3  & 100.0 & 0.0   & 37.6  & 47.1 & 14 & 5  \\
en-jv  & 59.3 & 55.3 & 48.0  & 39.2  & 52.7  & 0.0   & 0.0   & 28.0 & 14 & 5  \\
en-tl  & 55.4 & 53.5 & 35.0  & 70.5  & 26.7  & 0.0   & 38.8  & 34.2 & 14 & 6  \\
de-af  & 71.6 & 56.5 & 87.1  & 33.1  & 90.3  & 77.2  & 43.1  & 66.2 & 12 & 6  \\
de-ur  & 77.5 & 58.5 & 50.7  & 68.3  & 45.8  & 15.4  & 72.6  & 50.6 & 12 & 7  \\
de-sw  & 50.6 & 48.9 & 29.5  & 33.1  & 36.2  & 0.0   & 66.7  & 33.1 & 12 & 5  \\
de-te  & 71.2 & 55.7 & 45.6  & 29.4  & 5.2   & 0.0   & 56.5  & 27.3 & 12 & 7  \\
de-ta  & 76.3 & 57.6 & 43.0  & 56.7  & 48.7  & 0.0   & 81.3  & 45.9 & 12 & 7  \\
de-mn  & 74.7 & 59.1 & 44.4  & 68.3  & 42.8  & 0.0   & 61.8  & 43.4 & 12 & 5  \\
de-uz  & 62.8 & 55.1 & 48.3  & 91.9  & 27.8  & 0.0   & 81.1  & 49.8 & 12 & 6  \\
de-my  & 72.0 & 59.3 & 31.3  & 29.9  & 63.9  & 0.0   & 47.5  & 34.5 & 12 & 5  \\
de-jv  & 60.0 & 50.9 & 41.5  & 14.4  & 32.5  & 0.0   & 10.3  & 19.8 & 12 & 5  \\
de-tl  & 54.5 & 52.1 & 48.1  & 42.1  & 0.0   & 0.0   & 50.8  & 28.2 & 12 & 6  \\
zh-af  & 70.4 & 58.6 & 53.9  & 9.5   & 25.2  & 0.0   & 12.1  & 20.1 & 11 & 6  \\
zh-ur  & 75.1 & 62.8 & 59.0  & 43.5  & 36.3  & 0.0   & 82.6  & 44.3 & 11 & 7  \\
zh-sw  & 53.9 & 51.5 & 5.7   & 33.1  & 27.0  & 0.0   & 27.6  & 18.7 & 11 & 5  \\
zh-te  & 72.4 & 60.3 & 49.9  & 29.4  & 4.5   & 0.0   & 86.7  & 34.1 & 11 & 7  \\
zh-ta  & 73.0 & 61.8 & 19.0  & 56.7  & 16.8  & 0.0   & 40.5  & 26.6 & 11 & 7  \\
zh-mn  & 71.6 & 60.4 & 56.5  & 43.5  & 8.7   & 0.0   & 99.0  & 41.5 & 11 & 5  \\
zh-uz  & 62.5 & 54.9 & 49.0  & 69.3  & 26.2  & 0.0   & 87.2  & 46.3 & 11 & 6  \\
zh-my  & 69.6 & 59.3 & 42.5  & 71.8  & 32.7  & 37.8  & 95.7  & 56.1 & 11 & 5  \\
zh-jv  & 59.8 & 54.3 & 41.1  & 42.1  & 31.4  & 0.0   & 85.1  & 39.9 & 11 & 5  \\
zh-tl  & 54.7 & 52.4 & 44.7  & 14.4  & 6.9   & 0.0   & 83.4  & 29.9 & 11 & 6  \\
hi-af  & 78.2 & 59.0 & 55.4  & 50.1  & 30.8  & 14.3  & 52.3  & 40.6 & 7  & 6  \\
hi-ur  & 80.0 & 57.8 & 100.0 & 88.1  & 73.0  & 100.0 & 99.9  & 92.2 & 7  & 7  \\
hi-sw  & 50.7 & 50.5 & 27.4  & 24.6  & 24.9  & 0.0   & 66.9  & 28.8 & 7  & 5  \\
hi-te  & 72.7 & 58.4 & 74.7  & 74.4  & 67.2  & 0.0   & 100.0 & 63.3 & 7  & 7  \\
hi-ta  & 74.2 & 57.0 & 48.9  & 50.1  & 36.8  & 0.0   & 75.8  & 42.3 & 7  & 7  \\
hi-mn  & 74.6 & 57.7 & 57.9  & 61.3  & 31.2  & 0.0   & 89.4  & 48.0 & 7  & 5  \\
hi-uz  & 64.0 & 50.8 & 57.8  & 64.8  & 45.6  & 0.0   & 97.2  & 53.1 & 7  & 6  \\
hi-my  & 74.3 & 58.7 & 36.7  & 46.7  & 37.5  & 0.0   & 97.6  & 43.7 & 7  & 5  \\
hi-jv  & 59.4 & 48.7 & 21.2  & 0.0   & 13.6  & 0.0   & 79.6  & 22.9 & 7  & 5  \\
hi-tl  & 56.6 & 52.9 & 73.1  & 59.8  & 41.3  & 0.0   & 98.2  & 54.5 & 7  & 6  \\
ceb-af & 63.9 & 58.1 & 42.4  & 44.1  & 52.5  & 0.0   & 8.9   & 29.6 & 11 & 6  \\
ceb-ur & 68.7 & 57.1 & 29.3  & 84.3  & 22.5  & 0.0   & 62.9  & 39.8 & 11 & 7  \\
ceb-sw & 53.4 & 49.2 & 33.0  & 16.1  & 76.3  & 0.0   & 12.0  & 27.5 & 11 & 5  \\
ceb-te & 69.3 & 59.0 & 4.8   & 98.6  & 17.9  & 0.0   & 75.9  & 39.4 & 11 & 7  \\
ceb-ta & 66.3 & 55.8 & 22.4  & 72.1  & 63.0  & 0.0   & 16.6  & 34.8 & 11 & 7  \\
ceb-mn & 65.9 & 59.7 & 16.5  & 55.0  & 37.6  & 0.0   & 79.3  & 37.7 & 11 & 5  \\
ceb-uz & 56.2 & 52.6 & 26.2  & 61.3  & 17.9  & 0.0   & 60.6  & 33.2 & 11 & 6  \\
ceb-my & 64.8 & 56.3 & 3.0   & 43.5  & 57.7  & 0.0   & 88.1  & 38.4 & 11 & 5  \\
ceb-jv & 57.1 & 51.2 & 60.2  & 17.1  & 70.0  & 54.8  & 97.6  & 59.9 & 11 & 5  \\
ceb-tl & 53.0 & 56.2 & 0.0   & 82.7  & 50.0  & 0.0   & 76.2  & 41.8 & 11 & 6 \\
\bottomrule
\end{tabular}
\caption{Detailed data of 50 source-target language pairs used for correlation analysis of language similarity, source and target language pretraining data size with cross-lingual performance in unlabeled and labeled setup. Task performance is measured on Amazon review task with $k=1$.}
\tablabel{corr_analysis}
\end{table*}

\begin{table*}[t]
\scriptsize
\centering
\begin{tabular}{c|c|c|c|c|c|c|c|c|c|c|c|c|c}
\multicolumn{14}{c}{\textbf{Amazon Review}}\\
\hline
\multicolumn{2}{l|}{} & \textbf{en}   & \textbf{af}   & \textbf{ur}   &\textbf{ sw}   & \textbf{te }  & \textbf{ta}   & \textbf{mn}   & \textbf{uz}   & \textbf{my}   & \textbf{jv }  & \textbf{tl  } & \textbf{Avg}  \\ 
\hline
\multirow{4}{*}{UN}    & mBERT+pooling         & 57.8 & 54.4 & 54.9 & 52.4 & 53.5 & 54.8 & 51.1 & 49.3 & 52.4 & 56.1 & 52.1 & 53.1  \\ 
                       & mBERT+distiluse       & 63.1 & 60.1 & 61.0 & 46.1 & 50.1 & 50.0 & 59.9 & 55.2 & 56.7 & 57.2 & 50.1 & 54.7  \\ 
                       & mBERT+paraphrase      & \textbf{69.3} & 63.8 & 67.1 & 51.4 & 62.2 & 61.4 & 61.1 & 56.6 & 62.9 & 55.6 & 54.0 & 59.6  \\ 
                       & XLM-R+paraphrase      & 69.2 & \textbf{75.4} & \textbf{80.8} & \textbf{64.1 }& \textbf{71.0} & \textbf{70.4} & \textbf{69.7} & \textbf{68.2} & \textbf{70.4} & \textbf{63.8} & \textbf{66.6 }& \textbf{70.1 } \\ 
\hline
\multirow{4}{*}{LB}    & mBERT+pooling         & 65.6 & 56.8 & 57.0 & 51.8 & 53.8 & 53.1 & 52.7 & 51.2 & 52.5 & 53.5 & 53.2 & 53.6  \\ 
                       & mBERT+distiluse       & 80.4 & 76.0 & 80.0 & 51.2 & 48.9 & 50.0 & 77.9 & 57.7 & 70.7 & 60.5 & 55.4 & 62.8  \\ 
                       & mBERT+paraphrase      & \textbf{87.2} & \textbf{82.9} & \textbf{85.0} & 54.4 & \textbf{79.0} & \textbf{80.5} & \textbf{80.6} & 66.0 & \textbf{78.9 }& 61.5 & 60.2 & 72.9  \\ 
                       & XLM-R+paraphrase      & 77.6 & 81.7 & 82.2 & \textbf{64.0} & 74.2 & 73.9 & 75.1 & \textbf{70.6} & 76.4 & \textbf{66.3} & \textbf{66.1} & \textbf{73.0}  \\
\hline
\multicolumn{14}{c}{}\\
\multicolumn{14}{c}{\textbf{AG News}}\\
\hline
\multicolumn{2}{l|}{} & \textbf{en}   & \textbf{af}   & \textbf{ur}   &\textbf{ sw}   & \textbf{te }  & \textbf{ta}   & \textbf{mn}   & \textbf{uz}   & \textbf{my}   & \textbf{jv }  & \textbf{tl  } & \textbf{Avg}  \\
\hline
\multirow{4}{*}{UN}                                       & mBERT+pooling         & 37.9                    & 37.3                    & 34.8                    & 37.7                    & 32.9                    & 38.0                    & 36.0                    & 33.7                    & 37.4                    & 42.0                    & 38.8                    & 36.9                       \\ 
                                                          & mBERT+distiluse       & 43.3                    & 43.5                    & 38.8                    & 40.6                    & 25.4                    & 29.1                    & 39.7                    & 39.6                    & 42.7                    & 42.0                    & 42.9                    & 38.4                       \\ 
                                                          & mBERT+paraphrase      & 53.7                    & 52.8                    & 46.2                    & 46.5                    & 46.1                    & 42.8                    & 43.3                    & 44.3                    & 45.0                    & 51.0                    & 49.7                    & 46.7 \\ 
                                                          & XLM-R+paraphrase      & \textbf{62.7 }                   & \textbf{61.9  }                  & 58.9                    & \textbf{52.2   }                 & 58.1                    & \textbf{55.8 }                   & 55.6                    & \textbf{56.0}                    & \textbf{58.6  }                  & 59.2                    &\textbf{ 58.4    }                & \textbf{57.4}                       \\ 
\hline
\multirow{4}{*}{LB}                                       & mBERT+pooling         & 77.4                    & 68.2                    & 55.4                    & 58.5                    & 54.7                    & 52.1                    & 50.7                    & 54.6                    & 49.0                    & 66.7                    & 70.2                    & 58.0                       \\ 
                                                          & mBERT+distiluse       & 85.1                    & 82.0                    & 76.0                    & 65.5                    & 25.3                    & 28.7                    & 70.8                    & 64.4                    & 71.3                    & 77.8                    & 76.5                    & 63.8                       \\ 
                                                          & mBERT+paraphrase      & 74.9                    & 75.4                    & 68.1                    & 63.5                    & 68.2                    & 64.0                    & 62.8                    & 65.6                    & 64.8                    & 72.5                    & 71.4                    & 67.6  \\ 
                                             & XLM-R+paraphrase      & \textbf{83.8 }                   & \textbf{82.9   }                 &\textbf{ 78.8      }              &\textbf{ 70.4 }                   & \textbf{75.1  }                  & \textbf{71.7}                    & \textbf{72.7 }                   & \textbf{73.2  }                  &\textbf{ 77.4}                    & \textbf{79.2     }               & \textbf{79.0  }                  & \textbf{76.0}                       \\
\hline
\multicolumn{14}{c}{}\\
\multicolumn{14}{c}{\textbf{XNLI}}\\
\hline
\multicolumn{2}{l|}{} & \textbf{en}   & \textbf{af}   & \textbf{ur}   &\textbf{ sw}   & \textbf{te }  & \textbf{ta}   & \textbf{mn}   & \textbf{uz}   & \textbf{my}   & \textbf{jv }  & \textbf{tl  } & \textbf{Avg}  \\
\hline
\multirow{4}{*}{UN} & mBERT+pooling    & 34.7                    & 34.3                    & 34.4                    & 33.2                    & 33.9                    & 33.5                    & 34.3                    & 33.3                    & 33.3                    & 32.9                    & 32.7                    & 33.6                       \\ 
                    & mBERT+distiluse  & 32.9                    & 32.6                    & 33.4                    & \textbf{33.2 }                   & \textbf{36.1 }                   & 36.1                    & 33.8                    & 34.6                    & 31.9                    & \textbf{34.0}                    & 34.1                    & \textbf{34.0 }                      \\ 
                    & mBERT+paraphrase & 34.1                    & 32.9                    & \textbf{34.0 }                   & 33.0                    & 34.9                    & 34.7                    & 34.5                    & 34.5                    & 33.9                    & 31.4                    & 33.9                    & 33.7                       \\ 
                    & XLM-R+paraphrase & \textbf{35.5 }                   & \textbf{33.7 }                   & 34.0                    & 32.3                    & 35.0                    & \textbf{36.5}                    & \textbf{38.1  }                  & \textbf{34.7 }                   & \textbf{35.1}                    & 33.5                    & \textbf{34.1 }                   &\textbf{ 34.7 }                      \\ 
\hline
\multirow{4}{*}{LB} & mBERT+pooling    & 35.5                    & 34.1                    & 34.0                    & 35.3                    & 33.3                    & 34.1                    & 35.7                    & 32.8                    & 33.1                    & 33.5                    & 32.3                    & 33.8                       \\ 
                    & mBERT+distiluse  & 34.5                    & 35.6                    & 33.6                    & \textbf{35.1 }                   & 31.3                    & 31.4                    & 38.5                    & 35.6                    & 34.8                    & 35.7                    & 34.3                    & 34.6                       \\ 
                    & mBERT+paraphrase & \textbf{39.1}                    & \textbf{37.9 }                   & \textbf{37.8}                    & 33.5                    & \textbf{39.5   }                 & \textbf{39.4}                    & \textbf{39.1  }                  & 34.7                    & 36.9                    &\textbf{ 33.9 }                   &\textbf{ 35.7 }                   & \textbf{36.8  }                     \\ 
                    & XLM-R+paraphrase & 36.8                    & 35.7                    & 35.0                    & 32.8                    & 37.5                    & 37.5                    & 37.3                    & \textbf{36.7}                    & \textbf{37.5}                    & 32.8                    & 33.9                    & 35.7                       \\
\hline
\end{tabular}
\caption{Results of all languages using different combinations of retriever and MPLM for robustness analysis on Amazon review task ($k=5$), AG News tasks ($k=1$), and XNLI task ($k=3$), respectively.}
\tablabel{robust_results}
\end{table*}

\renewcommand{\tablename}{Table}
\begin{table*}[t]
	\small
	\centering
	\begin{tabular}{l|l|l|l|c}
		\hline 
		\textbf{Task} & \textbf{Dataset} & \textbf{Size} & \textbf{\#Label} & \textbf{Languages} \\
		\hline 
		Sentiment Analysis & Amazon Reviews & 1000 & 2 & af, ur, jv,\\ 
		\cline{1-4}
		Topic Categorization & AG News & 2000 & 4 &  ta, mn, uz,\\
		\cline{1-4}
		Sentence Pair Classification & XNLI & 1500 & 3 & tl, te, mn, sw \\
		\hline 
	\end{tabular}
	\caption{Overview of the test sets for the three tasks. Size refers to the number of samples for each LRL.}
	\tablabel{dataset}
\end{table*}

\end{document}